\documentclass[sigconf]{acmart}

\usepackage{booktabs} % For formal tables

\usepackage{graphicx} 
\usepackage{multirow}
\usepackage{multicol}
\usepackage{rotating}
\usepackage{amsmath}
\usepackage{amsthm}
\usepackage{amsfonts}
\usepackage{booktabs}
\usepackage{amssymb}
\usepackage{verbatim}
\usepackage{subfigure}
\usepackage{algorithm}  
\usepackage{algorithmic}
\usepackage{array}

\usepackage{url}
\usepackage{xspace}
\usepackage{enumitem}

\newcommand{\crossembed}{interaction embeddings\xspace}

\renewcommand{\shortauthors}{W. Zhang, B. Paudel, W. Zhang, A. Bernstein, and H. Chen}

\setcopyright{none}
\copyrightyear{2019}
\acmYear{2019}
\acmConference[WSDM '19]{The Twelfth ACM International Conference on Web Search and Data Mining}{February 11--15, 2019}{Melbourne, VIC, Australia}
\acmBooktitle{The Twelfth ACM International Conference on Web Search and Data Mining (WSDM '19), February 11--15, 2019, Melbourne, VIC, Australia}
\acmPrice{15.00}
\acmDOI{10.1145/3289600.3291014}
\acmISBN{978-1-4503-5940-5/19/02}

\begin{document}
\title{Interaction Embeddings for Prediction and Explanation in Knowledge Graphs}

\author{Wen Zhang}
\orcid{1234-5678-9012}
\affiliation{%
  \institution{College of Computer Science, Zhejiang University}
  %\institution{Zhejiang University}
  %\institution{Alibaba-Zhejiang University Joint Institute of Frontier Technologies}
  \country{China}
  %\streetaddress{38 Zheda Road}
  %\city{Hangzhou}
  %\state{Zhejiang}
  %\postcode{310007}
}
\email{wenzhang2015@zju.edu.cn}

\author{Bibek Paudel}
\affiliation{%
  \institution{Department of Informatics,\; University of Z\"urich}
  %\institution{University of Zurich}
  \country{Z\"urich, Switzerland}
}
\email{paudel@ifi.uzh.ch}

\author{Wei Zhang}
\affiliation{%
  \institution{Alibaba Group}
  \institution{AZFT Joint Lab of Knowledge Engine}
  \country{China}
 }
\email{lantu.zw@alibaba-inc.com}

\author{Abraham Bernstein}
\affiliation{%
  \institution{Department of Informatics,\; University of Z\"urich}
  %\institution{University of Zurich}
  \country{Z\"urich, Switzerland}
}
\email{bernstein@ifi.uzh.ch}

\author{Huajun Chen}
\authornote{Corresponding author.}
\affiliation{%
  \institution{College of Computer Science, Zhejiang University}
  \institution{AZFT Joint Lab of Knowledge Engine}
  %\country{China}
  %\streetaddress{38 Zheda Road}
  %\city{Hangzhou}
  %\state{Zhejiang}
  %\postcode{310007}
}
\email{huajunsir@zju.edu.cn}

\begin{abstract}
Knowledge graph embedding aims to learn distributed representations for entities and relations, and is proven to be effective in many applications. 
\emph{Crossover interactions} --- bi-directional effects between entities and relations --- help select related information when predicting a new triple, but haven't been formally discussed before. 
In this paper, we propose \emph{CrossE}, a novel knowledge graph embedding which explicitly simulates crossover interactions. 
It not only learns one general embedding for each entity and relation as most previous methods do, but also generates multiple triple specific embeddings for both of them, named \emph{\crossembed}. 
We evaluate embeddings on typical link prediction tasks and find that CrossE achieves state-of-the-art results on complex and more challenging datasets.
Furthermore, we evaluate embeddings from a new perspective --- giving explanations for predicted triples, which is important for real applications.
In this work, an explanation for a triple is regarded as a reliable closed-path between the head and the tail entity. 
Compared to other baselines, we show experimentally that CrossE, benefiting from \emph{\crossembed}, is more capable of generating reliable explanations to support its predictions.
\end{abstract}

\begin{comment}
\begin{CCSXML}
<ccs2012>
 <concept>
  <concept_id>10010520.10010553.10010562</concept_id>
  <concept_desc>Computer systems organization~Embedded systems</concept_desc>
  <concept_significance>500</concept_significance>
 </concept>
 <concept>
  <concept_id>10010520.10010575.10010755</concept_id>
  <concept_desc>Computer systems organization~Redundancy</concept_desc>
  <concept_significance>300</concept_significance>
 </concept>
 <concept>
  <concept_id>10010520.10010553.10010554</concept_id>
  <concept_desc>Computer systems organization~Robotics</concept_desc>
  <concept_significance>100</concept_significance>
 </concept>
 <concept>
  <concept_id>10003033.10003083.10003095</concept_id>
  <concept_desc>Networks~Network reliability</concept_desc>
  <concept_significance>100</concept_significance>
 </concept>
</ccs2012>
\end{CCSXML}

\ccsdesc[500]{Computer systems organization~Embedded systems}
\ccsdesc[300]{Computer systems organization~Redundancy}
\ccsdesc{Computer systems organization~Robotics}
\ccsdesc[100]{Networks~Network reliability}
\end{comment}
\vspace{-4mm}

\maketitle
%\input{samplebody-conf}
%!TEX root = crosse_wsdm.tex
\vspace{-2mm}
\section{Introduction}
\label{sec:introduction}
Knowledge graphs (KGs) like Yago~\cite{YAGO:conf/www/SuchanekKW07}, WordNet~\cite{WordNet:journals/cacm/Miller95}, and Freebase~\cite{Freebase:conf/sigmod/BollackerEPST08} have numerous facts in the form of (\emph{head entity}, \emph{relation}, \emph{tail entity}), or ( $h,r,t$ ) in short. They are useful resources for many AI tasks such as web search~\cite{WebSearch:conf/cikm/SzumlanskiG10} and question answering~\cite{QA:conf/ijcai/YinJLSLL16}.

\emph{Knowledge graph embedding} (KGE) learns distributed representations~\cite{RepresentationLearning/Hinton/1986} for entities and relations, called \emph{entity embeddings} and \emph{relation embeddings}. The embeddings are meant to preserve the information in a KG, and are represented as low-dimensional dense vectors or matrices in continuous vector spaces. 
Many KGEs, such as tensor factorization based RESCAL~\cite{RESCAL:conf/icml/NickelTK11}, translation-based TransE~\cite{TransE:conf/nips/BordesUGWY13}, neural tensor network NTN~\cite{NTN:conf/nips/SocherCMN13} and linear mapping method DistMult~\cite{DistMul:conf/iclr/2015}, have been proposed and are proven to be effective in many applications like knowledge graph completion, question answering and relation extraction.

Despite their success in modeling KGs, none of existing KGEs has formally discussed \textbf{crossover interactions}, bi-directional effects between entities and relations including \emph{interactions from relations to entities} and \emph{interactions from entities to relations}. 
Crossover interactions are quite common and helpful for related information selection, related information selection is necessary when predicting new triples because there are various information about each entity and relation in KGs. 

\begin{figure}[htp]
\vspace{-3mm}
\includegraphics[scale=0.38]{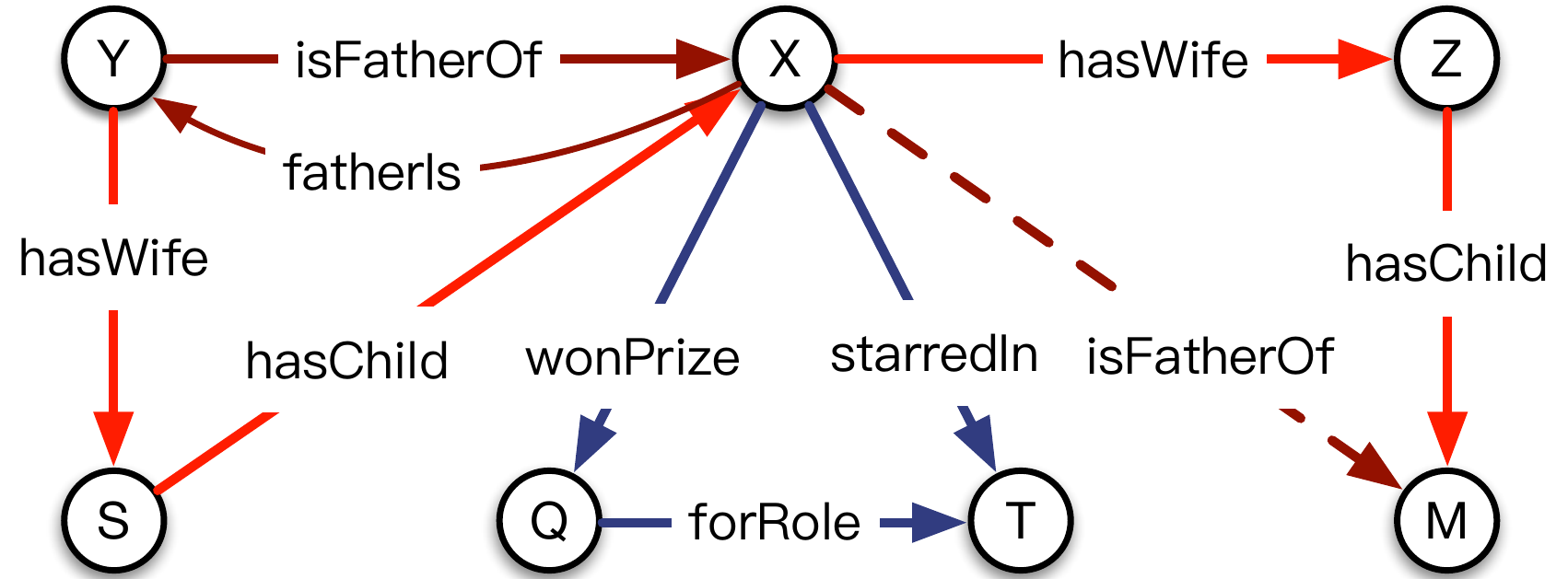}
\vspace{-2mm}
\caption{\small{A hypothetical knowledge graph. Nodes and edges represent entities and relations. Solid lines represent existing triples and dashed lines represent triples to be predicted.}}
\vspace{-3mm}
\label{story}
\end{figure}

We explain the notion of crossover interactions with a running example in Figure~\ref{story}. During the prediction of (\emph{X, isFatherOf, ?\;}), there are six triples about entity \emph{X}, but only four of them --- (\emph{X, hasWife, Z}), (\emph{X, fatherIs, Y}), (\emph{Y, isFatherOf, X}), (\emph{S, hasChild, X}) --- are related to this prediction, because they describe family relationships and will contribute to infer the father-child relationship. The other two triples describing the career relationship of \emph{X} do not provide valuable information for this task. In this way, the relation \emph{isFatherOf} affects the information of entities to be selected for inference. We refer this as \emph{interaction from relations to entities}.

In Figure~\ref{story}, there are two inference paths about the relation \emph{isFatherOf}, but only one of them --- ($X \xrightarrow{hasWife} Z \xrightarrow{hasChild} M $) --- is available for predicting (\emph{X, isFatherOf, ?}). In this way, the information of entity \emph{X} affects the inference path to be chosen for inference. We refer this as \emph{interactions from entities to relations}.

Considering \textbf{crossover interactions} in KGE, the embeddings for both entities and relations in specific triples should capture their interactions, and be different when involving different triples. However, most previous methods like TransE~\cite{TransE:conf/nips/BordesUGWY13} learn a general embedding, which is assumed to preserve all the information for each entity and relation. They ignore \emph{interactions} for both entities and relations. Some methods like TransH~\cite{TransH:conf/aaai/WangZFC14} and TransG~\cite{TransG:conf/acl/0005HZ16} learn either multiple entity or relation embeddings but not both. They ignore that crossover interactions are bi-directional and affect both entities and relations at the same time.

In this paper, we propose \textbf{CrossE}, a novel KGE which explicitly simulates \emph{crossover interactions}. 
It not only learns one general embedding for each entity and relation, but also generates multiple triple specific embeddings, called \emph{\crossembed}, for both of them. The \emph{\crossembed} are generated via a relation specific interaction matrix. 
Given an $(h, r, t)$, there are mainly four steps in CrossE: 
1) generate \emph{\crossembed} $\textbf{h}_I$ for head entity $h$; 
2) generate \emph{\crossembed} $\textbf{r}_I$ for relation $r$; 
3) combine {\crossembed}  $\textbf{h}_I$ and $\textbf{r}_I$ together; 
4) compare the similarity of combined embedding with tail entity embedding $\textbf{t}$.

We evaluate embeddings on canonical link prediction tasks. The experiment results show that CrossE achieves state-of-the-art results on complex and more challenging datasets and exhibits the effectiveness of modeling crossover interactions in a KGE.

Furthermore, we also propose an additional evaluation scheme for KGEs from the perspective of explaining their predictions.
Link prediction tasks only evaluate the accuracy of KGEs at predicting missing triples, while in real applications, explanations for predictions are valuable, as they will likely improve the reliability of predicted results. To the best of our knowledge, this is the first work to address both \textbf{link prediction and its explanation} of KGE.

The process of generating explanations for one triple $(h,r,t)$ is modeled as searching for reliable paths from $h$ to $t$ and similar structures to support path explanations. 
We evaluate the quality of explanations based on \emph{Recall} and \emph{Average Support}. \emph{Recall} reflects the coverage of triples that a KGE can generate explanations for and \emph{Average Support} reflects the reliability of the explanation.
Our evaluation on explanations show that CrossE, benefiting from \emph{interaction embeddings}, is more capable of giving reliable explanations than other methods, including TransE~\cite{TransE:conf/nips/BordesUGWY13} and ANALOGY~\cite{ANALOGY:conf/icml/LiuWY17}.

In summary, \textbf{our contributions} in this paper are the following:
\begin{itemize}[leftmargin=*, wide]
\item We propose CrossE, a new KGE which models crossover interactions of entities and relations by learning an interaction matrix.
\item We evaluate CrossE compared with various other KGEs on link prediction tasks with three benchmark datasets, and show that CrossE achieves state-of-the-art results on complex and more challenging datasets with a modest parameter size. 
\item We propose a new evaluation scheme for embeddings --- searching explanations for predictions, and show that CrossE is able to generate more reliable explanations than other methods. This suggests that {\crossembed} are better at capturing similarities between entities and relations in different contexts of triples.
\end{itemize}

\vspace{-1mm}
This paper is organized as follows. We review the literature in Section~\ref{sec:related_work}. We describe our model in Section~\ref{sec:model} and explanation scheme in Section~\ref{sec:explain}. We present the experimental results in Section~\ref{sec:experiments} before concluding in Section~\ref{sec:conclusion}.

%\input{motivationAnalysis}
%!TEX root=crosse_wsdm.tex
\vspace{-4mm}
\section{Related work}
\label{sec:related_work}
Knowledge graph embedding (KGE) aims to embed a knowledge graph into a continuous vector space and learns dense low dimensional representations for entities and relations.
Various types of KGE methods have been proposed and the majority of them learn the relationship between entities using training triples in a knowledge graph. Some methods also utilize extra information, such as logical rules, external text corpus and  hierarchical type information to improve the quality of embeddings. 
Since our work focuses on learning from triples in a knowledge graph without extra information, we mainly summarize those methods learning with triples and briefly summarize methods using extra information in the end.

Considering the requirement of multiple representations for entities and relations from crossover interactions, prior KGEs learning with triples can be classified into two classes: (1) methods learning a general embedding for each entity and relation, and (2) methods learning multiple representations for either of them.

\textbf{KGEs with general embeddings.}
Existing embedding methods with general embeddings all represent entities as low-dimensional vectors and relations as operations that combine the representation of head entity and tail entity. 
Most methods are proposed with different assumptions in vector space and model the knowledge graph from different perspectives.
The first translation-based method TransE~\cite{TransE:conf/nips/BordesUGWY13} regards relations as translations from head entities to tail entities and assumes that the relation-specific translation of head entity should be close to the tail entity vector. It represents each relation as a single vector.
RESCAL~\cite{RESCAL:conf/icml/NickelTK11} regards the knowledge graph as a multi-hot tensor and learns the entity vector representation and relation matrix representation via collective tensor factorization.
HOLE~\cite{HolE:conf/aaai/NickelRP16}, which is a compositional vector space model, utilizes interactions between different dimensions of embedding vectors and employs circular correlation to create compositional representations. 
RDF2Vec~\cite{RDF2Vec:conf/semweb/RistoskiP16} uses graph walks and Weisfeiler-Lehman subtree RDF kernels to generate entity sequence and regards the entity sequence as sequence of words in natural language, then follows word2vec to generate embeddings for entities but not relations.
NTN~\cite{NTN:conf/nips/SocherCMN13} represents each relation as a bilinear tensor operator followed by a linear matrix operator. 
ProjE~\cite{ProjE:conf/aaai/ShiW17} uses a simple but effective shared variable neural network.
DistMult~\cite{DistMul:conf/iclr/2015} learns embeddings from a bilinear objective where each relation is represented as a linear mapping matrix from head entities to tail entities. It successfully captures the compositional semantics of relations.
ComplEx~\cite{ComplEx:conf/icml/TrouillonWRGB16} makes use of complex valued embeddings to handle both symmetric and antisymmetric relations because the Hermitian dot product of real values is commutative while for complex values it is not commutative.
ANALOGY~\cite{ANALOGY:conf/icml/LiuWY17} is proposed from the analogical inference point of view and based on the linear mapping assumption.  It adds normality and commutatively constrains to matrices for relations so as to improve the capability of modeling analogical inference, and achieves the state-of-the-art results on link prediction task.

All these methods learn a general embedding for each entity and relation.  
They ignore crossover interactions between entities and relations when inferring a new triple in different scenarios.

\textbf{KGEs with multiple embeddings.}
Some KGEs learn multiple embeddings for entities or relations under various considerations.
Structured Embedding (SE)~\cite{SE:conf/aaai/BordesWCB11} assumes that the head entity and the tail entity in one triple should be close to each other in some subspace that depends on the relation. It represents each relation with two different matrices to transfer head entities and tail entities.
%%--ORC
ORC~\cite{ORC:conf/www/Zhang17} focuses on the one-relation-circle and proposes to learn two different representations for each entity, one as a head entity and the other as a tail entity.
%%---transH*
TransH~\cite{TransH:conf/aaai/WangZFC14} notices that TransE has trouble dealing with 1-N, N-1, and N-N relations.
It learns a specific hyperplane for each relation and represents the relation as a vector close to its hyperplane.
Entities are projected onto the relation hyperplane when involving a triple with this relation. 
%%---transR*
TransR~\cite{TransR:conf/aaai/LinLSLZ15} considers that various relations focus on different aspects of entities.
It represents different aspects by projecting entities from entity space to relation space and gets various relation specific embeddings for each entity.
CTransR~\cite{TransR:conf/aaai/LinLSLZ15} is an extension of TransR that considers 
correlations under each relation type by clustering diverse head-tail entity pairs into groups and learning distinct relation vectors for each group.
%%---transD*
TransD~\cite{TransD:conf/acl/JiHXL015} is a more fine-grained model which constructs a dynamic mapping matrix for each entity-relation pair considering the diversity of entities and relations simultaneously. 
%%---tranSparse
TranSparse~\cite{TranSparse:conf/aaai/JiLH016} is proposed to deal with the heterogeneity and imbalance of knowledge graph for relations and entities. It represents transfer matrices with adaptive sparse matrices and sparse degrees for transfer matrices are determined by the number of entities linked by this relation. 

These methods mostly consider the interaction from relations to entities  and learn multiple representations for entities. 
But they learn general embeddings for relations and ignore the interaction from entities to relations.

\textbf{KGEs that utilize extra information.}
Some KGEs learn embeddings utilizing not only training triples in a knowledge graph, but also extra information.
RTransE~\cite{RTransE:conf/emnlp/Garcia-DuranBU15}, PTransE~\cite{PTransE:conf/emnlp/LinLLSRL15} and CVSM~\cite{CVSM:conf/acl/NeelakantanRM15} utilize the path rules as additional constrains to improve embeddings. \cite{Rule:conf/ijcai/WangWG15} considers three types of physical rules and one logical rule, and formulates the inference in knowledge graph as an integer linear programming problem. \cite{TypeConstrain:conf/semweb/KrompassBT15} and TKRL~\cite{TKRL:conf/ijcai/XieLS16} propose that the hierarchical type information of entities is of great significance for the representation learning in knowledge graphs. \cite{TypeConstrain:conf/semweb/KrompassBT15} regards entity types as hard constraints in latent variable models for KGs and TKRL regards hierarchical types as projection matrices for entities.

%!TEX root = crosse_wsdm.tex
\vspace{-2mm}
\section{C\lowercase{ross}E: Model Description}
\label{sec:model}
In this section we provide details of our model CrossE.
Our model simulates crossover interactions between entities and relations by learning an \emph{interaction matrix} to generate multiple specific \crossembed.

In our method, each entity and relation is represented by multiple embeddings: 
(a) \emph{a general embedding}, which preserves high-level properties, and 
(b) multiple \emph{\crossembed}, which preserve specific properties as results of crossover interactions.
The \crossembed are obtained through Hadamard products between general embeddings and an \emph{interaction matrix} $\textbf{C}$.

We denote a knowledge graph as $\mathcal{K} = \{\mathcal{E}, \mathcal{R}, \mathcal{T}\}$, where $\mathcal{E}, \mathcal{R}$ and $ \mathcal{T}$ are the set of entities, relations and triples respectively.
The number of entities is $n_e$, the number of relations is $n_r$, and the dimension of embeddings is $d$.
Bold letters denote embeddings.
$\textbf{E} \in \mathbb{R}^{n_e \times d}$ is the matrix of general entity embeddings with each row representing an entity embedding.
Similarly, $\textbf{R} \in \mathbb{R}^{n_r \times d}$ is the matrix of general relation embeddings.
$\textbf{C} \in \mathbb{R}^{n_r \times d}$ is the \emph{interaction matrix} with each row related to a specific relation.

\begin{figure}[htbp] 
\vspace{-2mm}
\centering 
\includegraphics[scale=0.41]{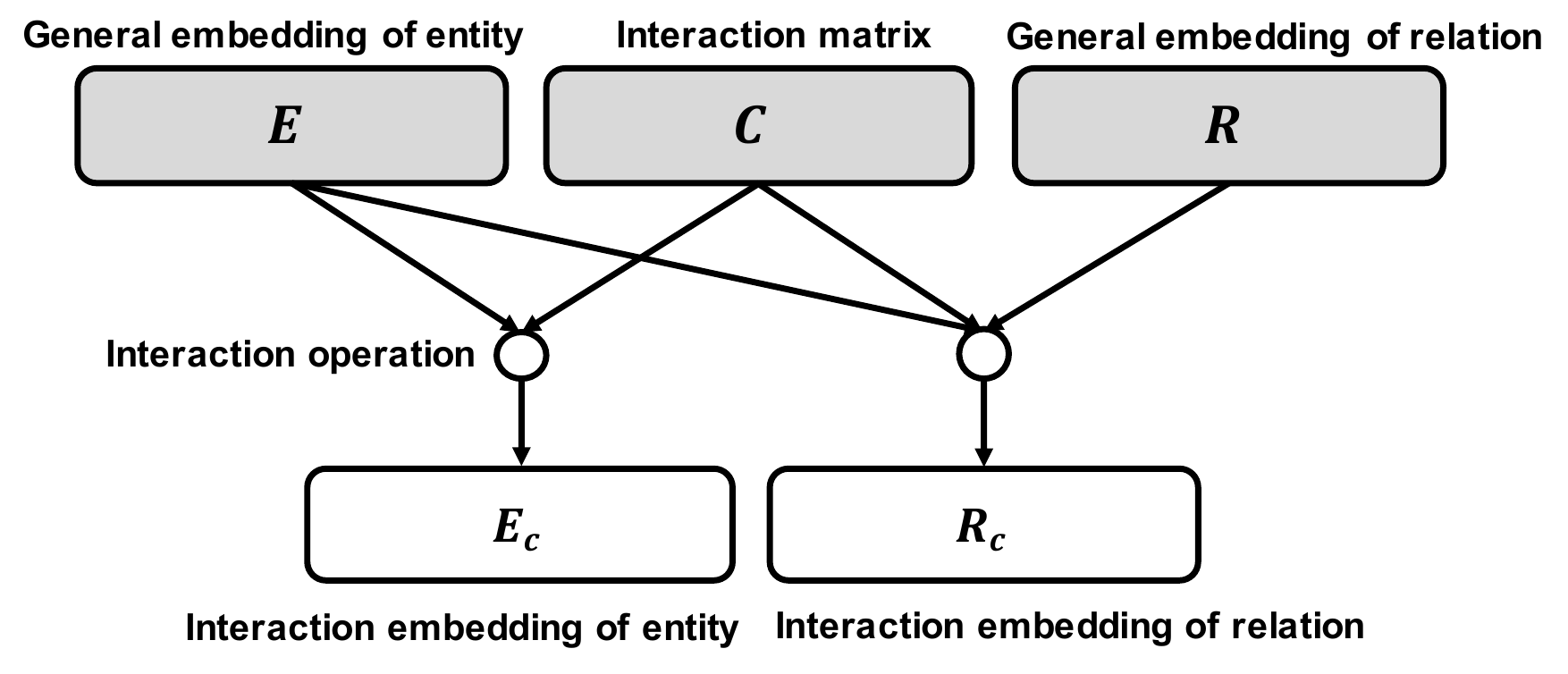}
\vspace{-4mm}
\caption{\small{Overview of CrossE. Crossover interactions between general embeddings ($\textbf{E}$ and $\textbf{R}$) and interaction matrix ($\textbf{C}$) resulting in interaction embeddings (unshaded boxes).}}
\label{cross affecting} 
\vspace{-5mm}
\end{figure}

The basic idea of CrossE is illustrated in Figure~\ref{cross affecting}.
The general embeddings ($\textbf{E}$ for entities and $\textbf{R}$ for relations) and interaction matrix $\textbf{C}$ are represented in the shaded boxes.
The interaction embeddings $\textbf{E}_c$ and $\textbf{R}_c$ are results of crossover interactions between entities and relations, which are fully specified by the interaction operation on general embeddings. 
Thus $\textbf{E}$, $\textbf{R}$, and $\textbf{C}$ are parameters that need to be learned, while $\textbf{E}_c$  and $\textbf{R}_c$ are not.

We now explain the score function and training objective of CrossE.
Head entity \emph{h}, tail entity \emph{t}, and relation \emph{r} correspond to high dimensional `one-hot' index vectors $\textbf{x}_h$, $\textbf{x}_t$ and $\textbf{x}_r$ respectively.
Learned general embeddings of \emph{h}, \emph{r} and \emph{t} are written as:
\begin{equation}
 \textbf{h} = \textbf{x}_h^\top \textbf{E}, \; \;  \textbf{t} = \textbf{x}_t ^ \top \textbf{E} , \; \; \textbf{r} = \textbf{x}_r^\top \textbf{R}
\end{equation}
In CrossE, we define a score function for each triple such that valid triples receive high scores and invalid triples receive low scores. The score function has four parts and we describe them next.

\begin{enumerate}[wide]
\item \textbf{Interaction Embedding for Entities.}
To simulate the effect from relations to head entities, we define the interaction operation applied to a head entity as:
\vspace{-1mm}
\begin{equation}
\label{eqref:entity_interaction}
\textbf{h}_I = \textbf{c}_r \circ \textbf{h}
\end{equation}
\vspace{-1mm}
, where $\circ$ denotes \emph{Hadamard product}, an element-wise operator that has been proved to be effective and widely used in previous methods, such as \cite{HolE:conf/aaai/NickelRP16} and \cite{ComplEx:conf/icml/TrouillonWRGB16}.
We call $\textbf{h}_I$ the interaction embedding of $h$.
Here, $\textbf{c}_r \in \mathbb{R}^{1 \times d}$ is a relation specific variable which we get from the interaction matrix $\textbf{C}$ as in~\eqref{eqref:relation_interaction_vector}. As $\textbf{c}_r$ depends on relation $r$, the number of interaction embeddings of $h$ are the same as relations.
\vspace{-1mm}
\begin{equation}
\label{eqref:relation_interaction_vector}
\textbf{c}_I = \textbf{x}_I ^ \top \textbf{C}
\vspace{-1mm}
\end{equation}

\item \textbf{Interaction Embedding for Relations.}
The second interaction operation is applied to relations so as to simulate the effects from head entities.
This interaction operation is defined in~\eqref{eqref:relation_interaction}. Similar to~\eqref{eqref:entity_interaction}, this is the Hadamard product of $\textbf{h}_I$ and \textbf{r} and we call $\textbf{r}_I$ the interaction embedding of $r$.
For each head entity, there is an interaction embedding of $r$.
\vspace{-1mm}
\begin{equation}
\label{eqref:relation_interaction}
\textbf{r}_I  = \textbf{h}_I \circ \textbf{r}
\vspace{-1mm}
\end{equation}
\vspace{-2mm}

\item \textbf{Combination Operator.}
The third step is to get the combined representation and we formulate it in a nonlinear way:
\begin{equation}
\textbf{q}_{hr} = tanh(\textbf{h}_I + \textbf{r}_I + \textbf{b})
\end{equation}
, where $\textbf{b} \in \mathbb{R}^{1 \times d}$ is a global bias vector. 
$tanh(z) = \frac{e^z - e^{-z}}{e^z + e^{-z}}$, in which the output is bounded from $-1$ to $1$. 
It is used to ensure the combined representation share the same distribution interval (both negative values and positive values) with the entity representation.

\item \textbf{Similarity Operator.}
The fourth step is to calculate the similarity between the combined representation $\textbf{q}_{hr}$ and the general tail entity representation $\textbf{t}$:
\vspace{-1mm}
\begin{equation}
\label{eq:similarity}
f(h,r,t) = s_{(h,r,t)} = \sigma(\textbf{q}_{hr} \textbf{t}^\top )
\vspace{-1mm}
\end{equation}
, where dot product is used to calculate the similarity and  $\sigma(z) = \frac{1}{1 + e ^{-z}}$ is a nonlinear function to constrain the score $s_{(h,r,t)} \in [ 0, 1]$.
\end{enumerate}

The overall score function is as follows:
\begin{equation}
f(h,r,t) = \sigma(tanh(\textbf{c}_r \circ \textbf{h} + \textbf{c}_r \circ \textbf{h} \circ \textbf{r} + \textbf{b}) \textbf{t}^\top )
\label{score function}
\end{equation}

To evaluate the effectiveness of the crossover interactions, we devise a simplified CrossE called CrossE$_S$ by removing the interaction embeddings and using only the general embeddings in the score function:
\vspace{-2mm}
\begin{equation}
f_S(h,r,t) = \sigma(tanh(\textbf{h} + \textbf{r} + \textbf{b})\textbf{t}^\top)
\vspace{-1mm}
\end{equation}

\textbf{Loss function.}
We formalize a log-likelihood loss function with negative sampling as the objective for training:
\begin{equation*}
\begin{split}
L(\mathcal{K}) = -  \sum_{(h,r,t) \in \mathcal{K}} & \ \ \sum_{x \in \mathcal{B}(h,r,t)}  \big[l(x)log(f(x))  \\
		& +(1-l(x))log(1-f(x)) \big] + \lambda\sum \| \theta \| _2 ^2
\end{split}
\end{equation*}
Here, $\mathcal{B}(h,r,t)$ is the bag of positive examples with label $1$ and negative examples with label $0$ generated for $(h,r,t)$.
The label of example $x$ is given by $l(x)$.
For $(h,r,t)$, positive examples are $(h,r,e) \in \mathcal{K}$ and negative examples are $(h,r,e) \notin \mathcal{K}$, where $e \in \mathcal{E}$.
The factor $\lambda$ controls the L2 regularization of model parameters $\theta  = \{ \textbf{E}, \textbf{R}, \textbf{C}, \textbf{b} \}$.
The training objective is to minimize $L(\mathcal{K})$, and we applied a gradient-based approach during training.

\textbf{Number of Parameters.}
The total number of parameters for CrossE is $(n_e + 2 n_r + 1) \times d$, as there are $n_e + n_r$ general embeddings, $n_r$ additional embeddings from the interaction matrix, and one bias term.
Note that interaction embeddings are fully specified by these parameters.
While predicting head entities, we model the task as the inverse of tail entity prediction, e.g., ($t, r^{-1}, h$).
In such cases, we need $2n_r$ more embeddings for inverse relations.
Since $n_r \ll n_e$ in most knowledge graphs, this does not add a lot of extra parameters.

\textbf{The main benefits of CrossE.} 
 Compared to existing KGEs, CrossE's benefits are as follows:
(1) For an entity and a relation, the representations used during a specific triple inference are \crossembed (not general embeddings), which simulate the selection of different information for different triple predictions.
(2) Multiple {\crossembed} for each entity and relation provide richer representations and generalization capability. We argue that they are capable of capturing different latent properties depending on the context of interactions. This is because each interaction embedding can select different similar entities and relations when involving different triples. 
(3) CrossE learns one general embedding for each entity and relation, and uses them to generate interaction embeddings. This results in much less extra parameters than learning multiple independent embeddings for each entity and relation.

%!TEX root = crosse_wsdm.tex
\vspace{-2mm}
\section{Explanations for Predictions}
\label{sec:explain}

In this section, we describe how we generate explanations for predicted triples. 
Explanations are valuable when KGEs are implemented in real applications, as they help improve the reliability of and people's trust on predicted results. The balance between achieving high prediction accuracy and giving explanations has already attracted research attention in other areas, such as recommender systems~\cite{OCuLaR:conf/icde/HeckelVPD17,TEM:conf/www/Wang0FNC18}.

We first introduce the motivation of our explanation scheme, followed by our embedding-based path-searching algorithm. 

\vspace{-2mm}
\subsection{Background}
Similar to inference chains and logical rules, meaningful paths from $h$ to $t$ can be regarded as explanations for a predicted triple $(h, r, t)$.
For example, in Figure~\ref{story}, the fact that X is father of M can be inferred by $(X, hasWife, Z)$ and $(Z, haschild, M)$.
\begin{equation}
X \xrightarrow{hasWife} Z \xrightarrow{hasChild} M  \Rightarrow X \xrightarrow{isFatherOf} M 
\label{path}
\end{equation}
The right-hand side of the implication "$\Rightarrow$" is called \emph{conclusion} and the left-hand side is \emph{premise}.
The \emph{premise} is an explanation for the \emph{conclusion}. In the above example, the path $X \xrightarrow{hasWife} Z \xrightarrow{hasChild} M$ is one explanation for the triple $(X, isFatherOf, M )$.

Searching paths between head entity and tail entity is the first step for giving explanations for a triple.
There are multiple works focusing on mining such paths as rules or features for prediction, e.g., AMIE+~\cite{AMIE+:journals/vldb/GalarragaTHS15} and PRA~\cite{PRA:conf/emnlp/LaoMC11}, in which the paths are searched and pruned based on random walks and statistical significance. 
An important aspect of efficient path searching is the volume of search space.
Selecting candidate start entities and relations is the key point of reducing search space. Good embeddings can be useful for candidate selection with less effects on path searching results, because they are supposed to capture the similarity semantics of entities and relations. 
Thus giving explanations for predicted triples, by searching for reliable paths based on embeddings, not only improves the reliability of predicted results, but also provides a new perspective of evaluating embedding qualities.

In this paper, the explanation reliability for triple $(h, r, t)$ is evaluated with the number of \emph{similar structures} in knowledge graph, on which the inferences are mainly based, as noted by~\cite{MLKG:journals/pieee/Nickel0TG16}. 
Similar structures contain same relations but different specific entities.

\begin{figure}[htbp] 
\vspace{-2mm}
\centering 
\includegraphics[scale=0.44]{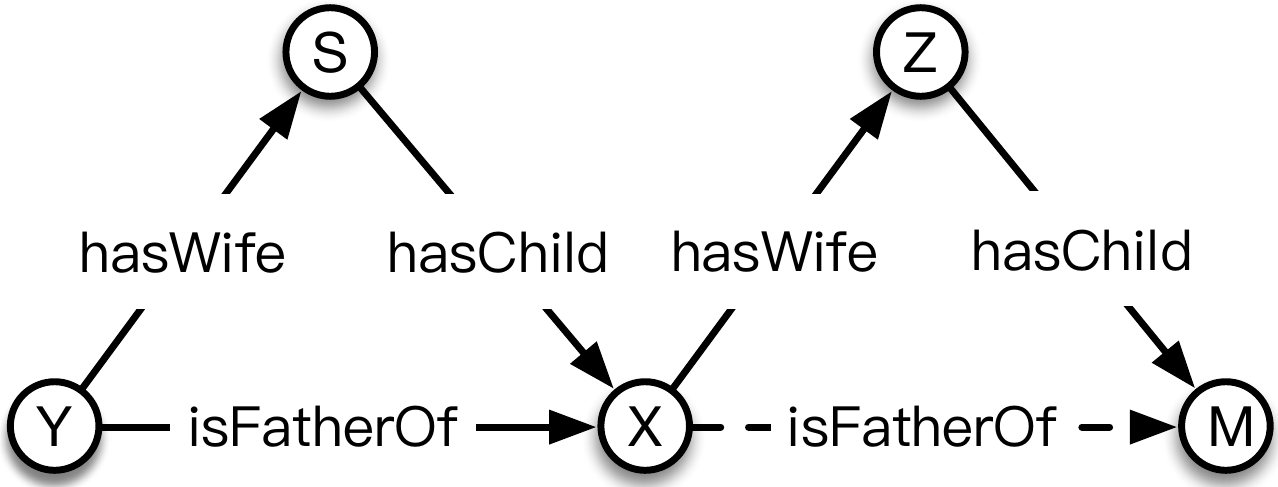}
\vspace{-2mm}
\caption{\small{Similar structures for the example subgraph in Figure \ref{story}.}}
\label{similarStructuresExample}
\vspace{-2mm}
\end{figure}

For example, the left and right subgraph in Figure~\ref{similarStructuresExample} have similar structures, as they both contain three entities $e_1, e_2, e_3$, a triple $e_1\xrightarrow{isFatherOf}e_3$, and a path $e_1\xrightarrow{hasWife}e_2\xrightarrow{hasChild}e_3$. 
Thus the left subgraph is a \emph{support} for $(X, isFatherOf, M)$ with path explanation $X\xrightarrow{hasWife}Z\xrightarrow{hasChild}M$ and vice versa. They support the reasonable existence and path explanation of each other.
In general, for an explanation, the more similar structure \emph{supports} it has, the more reliable it is.

\vspace{-4mm}
\subsection{Embedding-based explanation search}
During the embedding-based explanation search, we first select candidate entities and relations based on embedding similarity to reduce the search space before generating explanations for a triple $(h, r, t)$. The candidate selection is related to the quality of embeddings and will directly affect final explanations. We assume the similarity of vector embeddings are related to Euclidean distance and matrix embeddings are related to Frobenuis norm. Then based on selected candidates, we do exhaustive search for explanations which involves (1) searching for closed-paths from $h$ to $t$ as \emph{explanations} and (2) searching for similar structures of the explanations as \emph{supports}.
More specifically, there are four main steps:
\vspace{-1mm}
\begin{itemize}[wide]
\item \textbf{Step 1: Search for similar relations.} 
Output top $k_r$ relations similar to $r$, denoted by the set $\mathcal{S}_r = \{r_1, r_2, ... ,r_t \}$, as possible first steps for path search.
This step helps to prune unreasonable paths. 
For example, the path $h\xrightarrow{hasFriend}c\xrightarrow{likes}{b}$ doesn't indicate the relationship $h \xrightarrow{liveIn}{b}$, even if it may have a lot of supports, because $hasFriend$ is not relevent to the inference of $liveIn$.
To avoid such meaningless paths, the search is constrained to begin with relations similar to $r$, which are more likely to describe the same aspect of an entity.
\item \textbf{Step 2: Search for paths between $h$ and $t$.} 
Output a set of path $\mathcal{P} = \{p | (h,p,t) \in \mathcal{K} \}$. For simplicity, we consider paths including one or two relations as in~\cite{DistMul:conf/iclr/2015}. Thus there are six types of paths corresponding to six similar structures as shown in Table~\ref{analogyResults}. 
The six possible paths are: 
$p_{1} = \{ h \xrightarrow{r_s} t \}$, 
$p_{2} = \{ h \xleftarrow{r_s} t \}$, 
$p_{3} = \{  h \xleftarrow{r_s} e^\prime \xrightarrow{r^\prime} t \}$, 
$p_{4} = \{  h \xleftarrow{r_s} e^\prime \xleftarrow{r^\prime} t \}$, 
$p_{5} = \{  h \xrightarrow{r_s} e^\prime \xrightarrow{r^\prime} t \}$, 
$p_{6} = \{  h \xrightarrow{r_s} e^\prime \xleftarrow{r^\prime} t \}$.
Here, $e^\prime$ and $r^\prime$ denote any entity and relation in KG.
To search for paths with length two, we apply a bi-direction search strategy. 
Taking $p_5$ as an example,  we first find two entity sets $\mathcal{E}_1 = \{ e | (h, r_s, e) \in \mathcal{K}\}$ and $\mathcal{E}_2 = \{ e | (e, r^\prime, t) \in \mathcal{K}, r^\prime \in \mathcal{R}\}$. Then we get the paths via intersection entities of $\mathcal{E}_1$ and $\mathcal{E}_2$, $p = \{ (h \xrightarrow{r_s} e \xrightarrow{r^\prime} t ) |  (e, r ^\prime, t) \in \mathcal{K}, e \in \mathcal{E}_1 \cap \mathcal{E}_2 \}$. 
\item \textbf{Step 3: Search similar entities.} 
Find top $k_e$ similar entities  for $h$, denoted by the set $\mathcal{S}_h = \{h_1, h_2, ..., h_k \}$. 
Then check the corresponding tail entity $t_s$ of  $(h_s, r, ?)$ in the KG, where $h_s \in \mathcal{S}_h$. 
The tail checking results depend on the quality of selected similar entities. Therefore the more capable a KGE is in capturing similarity between entities, the more likely it is that $(h_s, r, t_s)$ exists.
\item \textbf{Step 4: Search for similar structures as supports.} Output supports for path $p \in \mathcal{P}$ from step 2 according to the similar entities from step 3. 
If $(h_s, p, t_s) \in \mathcal{K}$, path $p$ is an explanation for $(h,r,t)$ and $((h_s, p, r_s),$ $(h_s, r, t_s))$ is a support for this explanation.
We only regard paths with at least one support in knowledge graph as explanations. 
\end{itemize}

We summarize the process of embedding-based explanation search in Algorithm~\ref{algorithm:explanation}.
\begin{algorithm}
\small
\caption{Search for explanations for predicted triple $(h,r,t)$}
\label{algorithm:explanation}
\begin{algorithmic}[1]
\REQUIRE Knowledge graph $\mathcal{K}$, relation and entity embeddings $\textbf{R}$ and $\textbf{E}$
\ENSURE Explanations for $(h,r,t)$ and their supports.
\STATE Explanation set $E = \emptyset$, Support set $S = \emptyset $

\STATE Select the set $\mathcal{S}_r$ of top $k_r$ similar relations for $r$ 

\STATE Search the corresponding path set $\mathcal{P} = \{ p | (h, p, t) \in \mathcal{K}\}$ for each type: direct search for similar structure type 1 and type 2, and bidirectional search for type 3, type 4, type 5, and type 6.

\STATE Select the set $\mathcal{S}_h$ of top $k_e$ similar entities for $h$

\FOR{$p \in \mathcal{P}$}
\IF{$(h_s, p, t_s) \in \mathcal{K}$ and $h_s \in \mathcal{S}_h$}
\STATE $E \gets$ $E \cup p$ 
\STATE $S \gets$ $S \cup ((h_s, p, t_s),(h_s, r, t_s))$
\ENDIF
\ENDFOR
\end{algorithmic}
\end{algorithm}

In order to make the notion of explanations and supports more clear, in Table~\ref{analogyResults}, we give real examples with path explanations and their supports based on CrossE embeddings (Section~\ref{sec:explain_eval} provides details about the implementation).
For each type of path, we list a predicted triple with this kind of path explanation and one corresponding support from embedding-based explanation search results of CrossE. It is expected that the reliability of a predicted triple increases when end-users are also provided with explanations and their supports in the form of similar path structures.

%!TEX root = crosse_wsdm.tex
\vspace{-2mm}
\section{Experimental Evaluation}
\label{sec:experiments}
We use three popular benchmark datasets: WN18 and FB15k introduced in \cite{TransE:conf/nips/BordesUGWY13}, and FB15k-237 proposed by~\cite{Node-Leakf}.
They are either subset of WordNet~\cite{WordNet:journals/cacm/Miller95}, a large lexical knowledge base about English, or Freebase~\cite{Freebase:conf/sigmod/BollackerEPST08}, a huge knowledge graph describing general facts in the world. 
The details of these datasets are given in Table~\ref{tab:datasets}.
\vspace{-3mm}
\begin{table}[!h]
\small
\centering
  \begin{tabular}{ c |c  c  c  c c}
    \toprule
      Dataset & $\mid\mathcal{E}\mid$ & $\mid\mathcal{R}\mid$ & $\mid$Train Set$\mid$ & $\mid$Validation Set$\mid$  & $\mid$Test Set$\mid$\\
      \midrule
      WN18   & 40,943 & 18 & 141,442 & 5,000 & 5,000 \\
      FB15k  & 14,951 & 1,345 & 483,142 & 50,000 & 59,071 \\
      FB5k-237 & 14,541 & 237 & 272,115 & 17,535 & 20,466 \\
    \bottomrule
  \end{tabular}
  \caption{\small{Datasets statistics.}}
  \vspace{-6mm}
\label{tab:datasets}
\end{table}
\vspace{-2mm}
\begin{table*}[htb]
\footnotesize
\centering
\begin{tabular}{|c |c |c| l l l|}
\hline
\multicolumn{3}{ |c|}{ } & \textbf{Head entity} & \textbf{Relation} & \textbf{Tail entity} \\
\hline
% =====================================================
\multirow{4}{*}{\rotatebox{90}{\textbf{Type 1}}} 
&\multirow{4}{*}{ \includegraphics[scale=0.355]{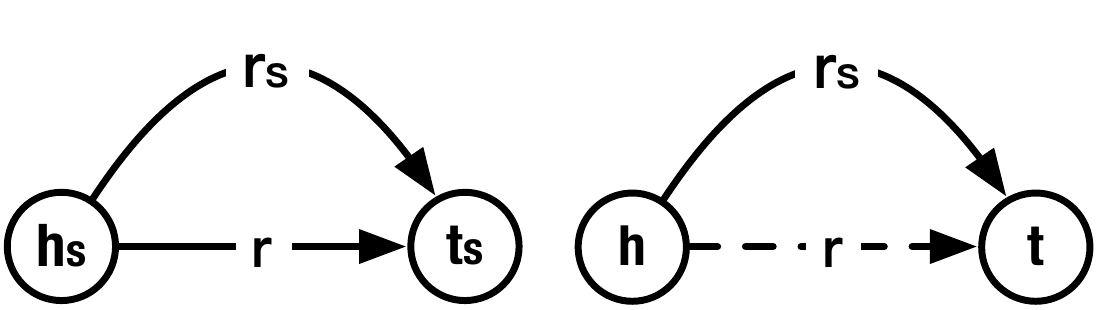}} 
& \textbf{Predicted triple} 
& Mel Gibson & \texttt{award nominations} &  Best Director \\

\cline{3-6}
&
& \textbf{Explanation} 
& Mel Gibson & \texttt{awards won} & Best Director \\

\cline{3-6}
&
& \multirow{2}{*}{\textbf{Support}} 
& Vangelis & \texttt{award nominations} & Best Original Musical \\

\cline{4-6}
&
& 
& Vangelis &\texttt{awards won} & Best Original Musical \\
\hline

% =====================================================
\multirow{4}{*}{\rotatebox{90}{\textbf{Type 2}}}
&\multirow{4}{*}{\includegraphics[scale=0.355]{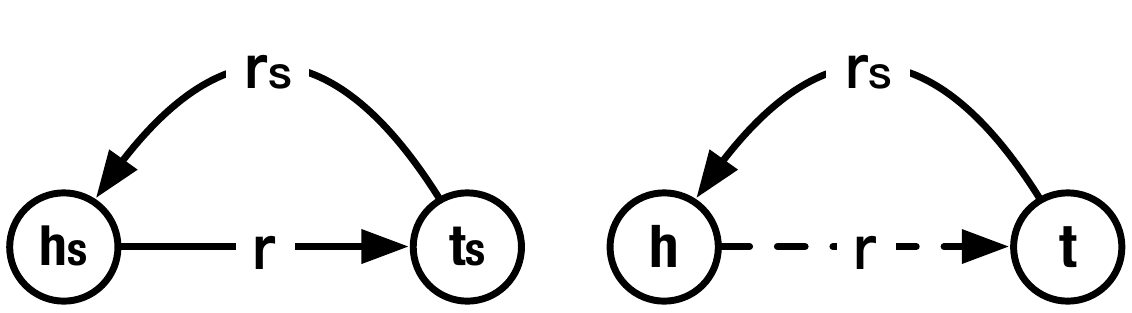}}
& \textbf{Predicted triple }
& Aretha Franklin &\texttt{influenced} &Kings of Leon \\

\cline{3-6}
&
& \textbf{Explanation}
& Kings of Leon &\texttt{influenced by} &Aretha Franklin \\

\cline{3-6}
&
& \multirow{2}{*}{\textbf{Support} }
& Michael Jackson &\texttt{influenced} &Lady Gaga \\

\cline{4-6}
&
&
& Lady Gaga &\texttt{influenced by} & Michael Jackson \\

\hline

% =====================================================
\multirow{6}{*}{\rotatebox{90}{\textbf{Type 3}}}
&\multirow{6}{*}{\includegraphics[scale=0.355]{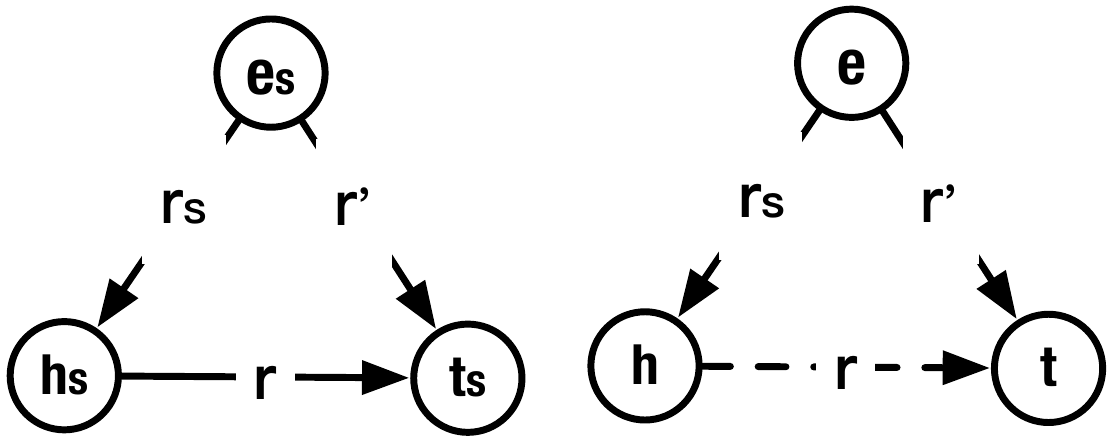} }
& \textbf{Predicted triple} 
& Cayuga County &\texttt{containedby} &New York \\

\cline{3-6}
&
& \multirow{2}{*}{\textbf{Explanation}}
& Auburn &\texttt{capital of} &Cayuga County \\

\cline{4-6}
&
& 
& Auburn &\texttt{containedby} &New York \\

\cline{3-6}
&
&\multirow{3}{*}{\textbf{Support}}
& Onondaga County &\texttt{containedby} &New York \\

\cline{4-6}
&
& 
& Syracuse  &\texttt{capital of} &Onondaga County \\
\cline{4-6}
&
& 
& Syracuse &\texttt{containedby} &New York \\

\hline

% =====================================================

\multirow{6}{*}{\rotatebox{90}{\textbf{Type 4}}}
&\multirow{6}{*}{\includegraphics[scale=0.355]{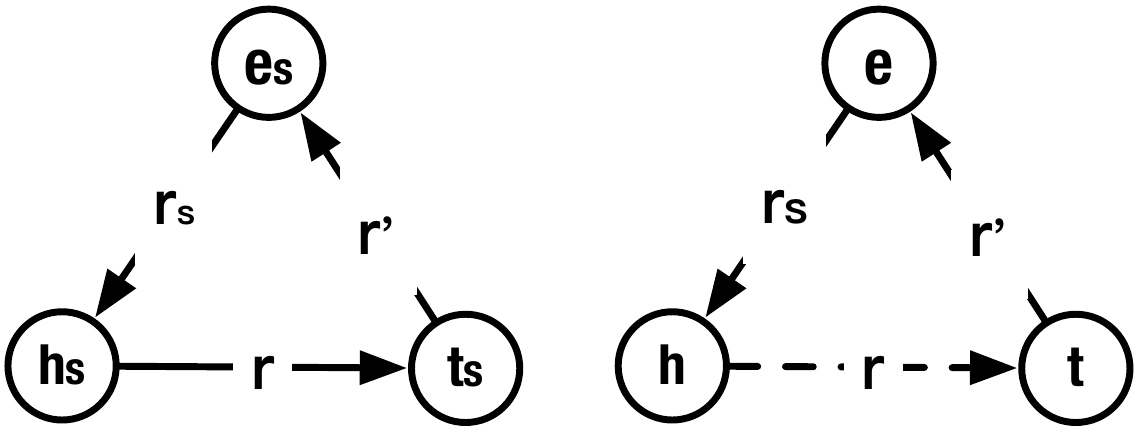} } 
& \textbf{Predicted triple }
& South Carolina &\texttt{country} &USA \\

\cline{3-6}
&
& \multirow{2}{*}{\textbf{Explanation}}
& Columbia &\texttt{state} &South Carolina \\

\cline{4-6}
&
& 
& United States of America &\texttt{contains} &Columbia \\

\cline{3-6}
&
&\multirow{3}{*}{\textbf{Support}}
& Mississippi &\texttt{country} &USA \\

\cline{4-6}
&
& 
& Jackson &\texttt{state} &Mississippi \\

\cline{4-6}
&
& 
& United States of America &\texttt{contains} &Jackson \\

\hline

% =====================================================
\multirow{6}{*}{\rotatebox{90}{\textbf{Type 5}}}
&\multirow{6}{*}{\includegraphics[scale=0.355]{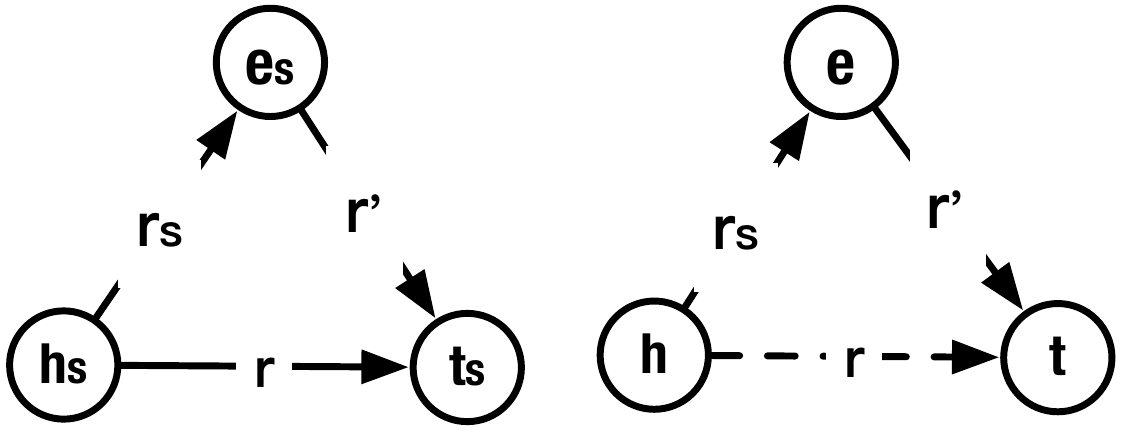} } 
& \textbf{Predicted triple }
& World War I &\texttt{entity involved} &German Empire \\

\cline{3-6}
&
& \multirow{2}{*}{\textbf{Explanation}}
&  World War I &\texttt{commanders} &Erich Ludendorff \\
\cline{4-6}
&
& 
& Erich Ludendorff &\texttt{military commands} &German Empire\\
\cline{3-6}
&
&\multirow{3}{*}{\textbf{Support}}
& Falklands War &\texttt{entity involved} &United Kingdom \\

\cline{4-6}
&
& 
& Falklands War &\texttt{commanders} &Margaret Thatcher \\

\cline{4-6}
&
& 
& Margaret Thatcher &\texttt{military commands} &United Kingdom \\

\hline

% =====================================================
\multirow{6}{*}{\rotatebox{90}{\textbf{Type 6}}}
& \multirow{6}{*}{\includegraphics[scale=0.355]{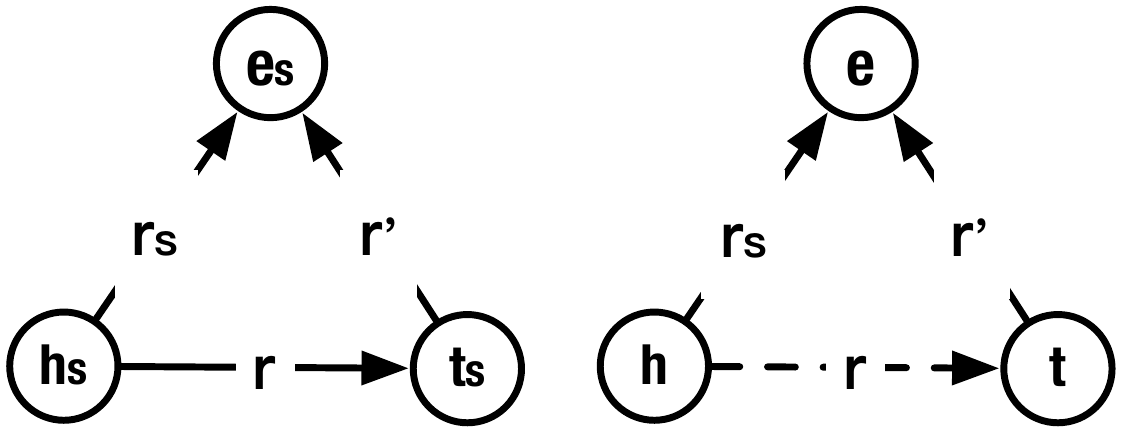}} 
& \textbf{Predicted triple }
& Northwestern University &\texttt{major field of study} &Computer Science \\

\cline{3-6}
&
& \multirow{2}{*}{\textbf{Explanation}}
& Northwestern University &\texttt{specialization} &Artificial intelligence \\

\cline{4-6}
&
& 
& Computer Science &\texttt{specialization} &Artificial intelligence \\

\cline{3-6}
&
&\multirow{3}{*}{\textbf{Support}}
&Stockholm University &\texttt{major field of study} &Philosophy \\

\cline{4-6}
&
& 
& Stockholm University &\texttt{specialization} &Political\_philosoph \\

\cline{4-6}
&
& 
& Philosophy &\texttt{specialization} &Political\_philosoph \\

\hline

 \end{tabular}
\caption{\small{Six similar structures with real examples of explanations and supports for predictions made by CrossE.}}
 \label{analogyResults}
 \vspace{-7mm}
\end{table*}

We now explain our evaluation on two main tasks: (a) link prediction, and (b) generating explanations for predicted triples.
\vspace{-2mm}
\subsection{Evaluation \uppercase\expandafter{\romannumeral1}: Link Prediction}
In this section, we evaluate embeddings on canonical link prediction tasks, which contain two subtasks, one is tail entity prediction $(h,r,?)$ and the other is head entity prediction $(?, r, t)$.

\subsubsection{Evaluation Metrics} 
The link prediction evaluation follows the same protocol as previous works, in which all entities in the dataset are candidate predictions.
During head entity prediction for $(h,r,t)$, we replace $h$ with each entity in the dataset and calculate their scores according to the score function~\eqref{score function}.
Then we rank the scores in descending order and regard the rank of $h$ as head entity prediction result.
The tail entity prediction is done similarly.

Aggregating head and tail entity prediction ranks of all test triples, there are three evaluation metrics: 
\emph{Mean Rank(MR)}, \emph{Mean Reciprocal Rank(MRR)} and \emph{Hit@N}. \emph{Hit@N} is the proportion of ranking scores within $N$ of all test triples.
\emph{MRR} is similar to \emph{MR} but more immune to extremely bad cases.  
Similar to most recent works, we evaluate CrossE on \emph{MRR} and \emph{Hit@1, Hit@3} and \emph{Hit@10}. We express both \emph{MRR} and \emph{Hit@k} scores as percentages. 

We also apply \emph{filter} and \emph{raw} settings. In \emph{filter} setting, we filter all candidate triples in train, test or validation datasets before ranking, as they are not negative triples. \emph{Raw} is the setting without filtering.
\vspace{-4mm}

\subsubsection{Implementation Details}
$\textbf{E},\textbf{R}$ and $\textbf{C}$ are initialized from the uniform distribution $U[-\frac{6}{\sqrt{d}},\frac{6}{\sqrt{d}}]$ as suggested in \cite{Init:journals/jmlr/GlorotB10}. $\textbf{b}$ is initialized to zero. 
The positive samples for training triple $(h,r,t)$ are generated by retrieving all $(h,r,e) \in \mathcal{K}_{train}$. Negative triples are generated randomly by sampling $n$ entities such that $(h,r,e) \notin \mathcal{K}_{train}$. 
The head entity prediction $(?, r,t)$ is transformed to $(t, r^{-1}, ?)$ in which $r^{-1}$ is the inverse relation of $r$. We generate inverse triples $(t,r^{-1},h)$ for each $(h,r,t)$ during training, as done in \cite{ CVSM:conf/acl/NeelakantanRM15, PTransE:conf/emnlp/LinLLSRL15}.

We implement our model using TensorFlow with Adam optimizer~\cite{Adam:journals/corr/KingmaB14} and dropout~\cite{dropout:journal/JMLR/2014} rate $0.5$ applied to similarity operator in~\eqref{eq:similarity}. 
The maximum training iteration is set to $500$ for all datasets. 
The configurations for the results of CrossE are as follows: number of negative examples $n = 50$, learning rate $r = 0.01$, dimension $d=100$, regularizer parameter $\lambda = 10^{-4}$ and batch-size $B = 2048$ for  WN18;  $n = 50$, $r = 0.01$, $d=300$, $\lambda = 10^{-6}$ and $B = 4000$ for  FB15K; $n = 50$, $r = 0.01$, $d=100$, $\lambda = 10^{-5}$ and $B = 4000$ for FB15k-237. CrossE$_S$ is trained with the same parameters. 

\subsubsection{Link Prediction Results}
\begin{table}[!htbp]
\small
\centering
  \begin{tabular}{ l| c |c| c  }
    \toprule
              & \#parameters &   WN18    &   FB15k \\
    \hline
    Unstructured \cite{Unstructed/SME:journals/ml/BordesGWB14}  & $n_e d$   &   38.2    &   6.3 \\
    RESCAL\cite{RESCAL:conf/icml/NickelTK11}    &  $n_e d + n_r d^2$       &   52.8    &   44.1    \\
    NTN  \cite{NTN:conf/nips/SocherCMN13}       & $n_e d + n_r(sd^2 + 2sd + 2s)$        &   66.1    &   41.4    \\
    SE  \cite{SE:conf/aaai/BordesWCB11}    &  $n_e d + 2n_r d$    &   80.5    &   39.8    \\
    LFM  \cite{LFM:conf/nips/JenattonRBO12} &   $n_e d + n_r d^2$      &   81.6    &   33.1    \\
    TransH\cite{TransH:conf/aaai/WangZFC14}  &  $n_e d + 2n_r d$       &   86.7    &   64.4    \\
    TransE\cite{TransE:conf/nips/BordesUGWY13} & $n_e d + n_r d$        &   89.2    &   47.1    \\
    TransR\cite{TransR:conf/aaai/LinLSLZ15}    & $n_e d + n_r d^2 + n_r d$     &   92.0    &   68.4    \\
    RTransE  \cite{RTransE:conf/emnlp/Garcia-DuranBU15}  & $n_e d + n_r d$  &   -       &   76.2    \\
    TransD\cite{TransD:conf/acl/JiHXL015}  &   $2n_e d + 2n_r d$     &   92.2    &   77.3     \\
    CTransR \cite{TransR:conf/aaai/LinLSLZ15} &  $n_e d + n_r d^2$    &   92.3    &   70.3    \\
    KG2E    \cite{KG2E:conf/cikm/HeLJ015}  &  $2n_e d + 2n_r d$    &   93.2    &   74.0    \\
    STransE  \cite{STransE:conf/naacl/NguyenSQJ16} & $n_e d + n_r d + 2 n_r d^2 $     &   93.4    &   79.7    \\
    DistMult \cite{DistMul:conf/iclr/2015}  & $n_e d + n_r d^2$    &   93.6    &   78.3    \\
    TranSparse \cite{TranSparse:conf/aaai/JiLH016} & $n_e d + n_r d + (1-\theta)n_r d^2 $ &   93.9   &    79.9    \\
    PTransE-MUL \cite{PTransE:conf/emnlp/LinLLSRL15} & $n_e d + n_r d $  &   -       &   77.7    \\
    PTransE-RNN \cite{PTransE:conf/emnlp/LinLLSRL15} & $n_e d + n_r d + d^2$  &   -       &   82.2    \\
    PTransE-ADD \cite{PTransE:conf/emnlp/LinLLSRL15} & $n_e d + n_r d$  &   -       &   84.6    \\
    ComplEx \cite{ComplEx:conf/icml/TrouillonWRGB16} & $2n_e d + 2n_e d$      &   94.7    &   84.0    \\
    ANALOGY  \cite{ANALOGY:conf/icml/LiuWY17}  & $n_e d + n_r d$    &   94.7    &   85.4    \\
    HOlE \cite{HolE:conf/aaai/NickelRP16}   &  $n_e d + n_r d$      &   94.9    &   73.9    \\
    \hline
    CrossE    &  $n_e d + 2n_r d + d$    &   \textbf{\underline{95.0}}    &  \underline{\textbf{87.5}}$^\dagger$    \\
    CrossE$_{S}$  & $n_e d + n_r d + d$   & 87.3 &  72.7\\
    \bottomrule
  \end{tabular}
\caption{\small{HIT@10(filter) results of 22 KGEs %(including CrossE) 
on WN18 and FB15k. 
"-" indicates missing results from original paper. Boldface scores are the best results among all methods. 
Underlined scores are the better ones between CrossE and CrossE$_{S}$. In \emph{parameters} column, $n_e$ and $n_r$ denote number of entities and relations respectively, 
$d$ is the embedding dimension, $s$ is the number of hidden nodes of a neural network, $\theta$ is the sparse degree of matrix.
}}
\label{link prediction results}
\vspace{-8mm}
\end{table}

In Table~\ref{link prediction results}, we show the results of CrossE and 21 baselines with their published results of \emph{Hit@10(filter)} on WN18 and FB15k from the original papers\footnote{We follow the established practice in the KGE literature to compare link-prediction performance with published results on the same benchmarks.}. This is the most applied evaluation metric on the most commonly used datasets in prior works, as we want to compare CrossE with as many baselines as possible.
For fair comparison, models utilizing external information, such as text, are not considered as baselines.

All CrossE results that are significantly different from the second-best results are marked with $\dagger$. We used one-sample proportion test at the 5\% p-value level for testing the statistical significances\footnote{Similar to~\cite{ANALOGY:conf/icml/LiuWY17}, we conducted the
proportion tests on the Hit@k scores but not on MRR. Proportion tests cannot be applied to non-proportional scores such as MRR.}.

In Table~\ref{LPdetail}, we compare CrossE with seven baseline methods whose \emph{MRR, Hit@1} and \emph{Hit@3} results are available.  
The results of TransE, DistMult and ComplEx are from \cite{ComplEx:conf/icml/TrouillonWRGB16}, RESCAL and HOLE from \cite{HolE:conf/aaai/NickelRP16}, ANALOGY from \cite{ANALOGY:conf/icml/LiuWY17} and R-GCN from \cite{R-GCN:conf/eswc}.

\begin{table}[htb]
\vspace{-2mm}
\small
\centering
  \begin{tabular}{ l |>{\hfil}p{1.2cm}>{\hfil}   p{0.6cm} p{0.5cm}  | >{\hfil}p{1.2cm}>{\hfil} p{0.6cm}  p{0.5cm} }
    \toprule

    \multirow{3}{*}{}               & \multicolumn{3}{c|}{WN18} &  \multicolumn{3}{c}{FB15k}  \\
    \cline{2-7}
                                    & MRR       &\multicolumn{2}{|c|}{Hit@}   &MRR       &\multicolumn{2}{|c}{Hit@}   \\
    \cline{2-7}                                                    & \multicolumn{1}{c|}{filter/raw}        &1          &3 &\multicolumn{1}{c|}{filter/raw} &1 &3 \\
    \hline
    \small{RESCAL\cite{RESCAL:conf/icml/NickelTK11}}            
    & 89.0 / 60.3       &84.2       &90.4       &35.4 /18.9     &23.5       &40.9       \\
    \small{TransE\cite{TransE:conf/nips/BordesUGWY13}}          
    &45.5 / 33.5        &8.9        &82.3       &38.0 / 22.1        &23.1       &47.2       \\
    \small{DistMult\cite{DistMul:conf/iclr/2015}}               
    &82.2 / 53.2        &72.8       &91.4       &65.4 / 24.2        &54.6       &73.3       \\
   \small{HOlE\cite{HolE:conf/aaai/NickelRP16}}                
   &93.8 / 61.6        &93.0       &\textbf{94.5}      &52.4 / 23.2        &40.2       &61.3       \\
    \small{ComplEx\cite{ComplEx:conf/icml/TrouillonWRGB16}}    
    &94.1 / 58.7        &93.6       &\textbf{94.5}      &69.2 / 24.2        &59.9       &75.9   \\
    \footnotesize{ANALOGY\cite{ANALOGY:conf/icml/LiuWY17}}             
    &\textbf{94.2}/\textbf{65.7}      &\footnotesize{\textbf{93.9}}      &94.4       &72.5 / 25.3        &\textbf{64.6}      &78.5       \\
    \footnotesize{R-GCN~\cite{R-GCN:conf/eswc}}                        
    &81.9 / 56.1        &69.7       &92.9       &69.6 / 26.2      &60.1       &76.0       \\              
    \hline
    \small{CrossE}                                          &   \underline{83.0} / \underline{57.0}     &   \underline{74.1}        &   \underline{93.1}        &\underline{\textbf{72.8}}/\underline{\textbf{26.7}}  &\underline{63.4}       &\underline{\textbf{80.2}}$^\dagger$      \\
    \small{CrossE$_{S}$}                                    & 46.9 / 39.6 & 21.7 & 70.6    &46.4 / 25.4&28.4&61.9 \\
    \bottomrule

\end{tabular}
\caption{\small{Link prediction results on WN18 and FB15k.}}
\label{LPdetail}
\vspace{-6mm}
\end{table}

In Table~\ref{LPdetail2}, we separately show the link prediction results of CrossE on FB15k-237, a recently proposed more challenging dataset.  
We gather as many baselines as possible for FB15k-237 and list them in Table~\ref{LPdetail2}. The results of DistMult, Node+LinkFeat and Neural LP are from~\cite{NeurlLP:conf/nips/YangYC17}, and  R-GCN and R-GCN+ from the original paper~\cite{R-GCN:conf/eswc}. For ComplEx and ANALOGY, we use the published code\footnote{https://github.com/quark0/ANALOGY} in~\cite{ANALOGY:conf/icml/LiuWY17} to decide the best parameters based on grid search among embedding dimension $d\in{\{100, 200\}}$ and regularizer weight $\lambda \in{\{10 ^{-1}, 10^{-2}, 10^{-3} \}}$ with six negative samples as in the ANALOGY paper~\cite{ANALOGY:conf/icml/LiuWY17}~\footnote{Our parameter search includes the same range of values used in the original papers, and the best parameters obtained for FB15.}. The parameters used for ComplEx and ANALOGY are $d=100$, and $ \lambda = 10^{-1}$.
\begin{table}[htb]
\small
\centering
  \begin{tabular}{ l |c c  c c c}
    \toprule

    \multirow{3}{*}{}							 &  \multicolumn{5}{c}{FB15k-237} \\
    \cline{2-6}
    								&MRR		&MRR		&Hit@1		&Hit@3	& Hit@10	\\
    
    								&(raw)	&(filter)	&(filter) &(filter) &(filter)	\\
    \hline
	
	DistMult\cite{DistMul:conf/iclr/2015} 
    & - & 25.0 & - & -  & 40.8 \\
	Node+LinkFeat \cite{Node-Leakf} 
    & - & 23.0 &-  &- & 34.7 \\
	Neural LP\cite{NeurlLP:conf/nips/YangYC17}
    & - & 24.0 &-  &- & 36.2 \\ 
    R-GCN \cite{R-GCN:conf/eswc}					
    &15.8		&24.8 		&15.3 & 25.8 & 41.4		\\
    R-GCN+ \cite{R-GCN:conf/eswc}  						
    &15.6		&24.9 		&15.1 & 26.4 & 41.7		\\
    ComplEx \cite{ComplEx:conf/icml/TrouillonWRGB16}
    & 12.0 & 22.1 & 13.2 & 24.4 & 40.8  \\
    ANALOGY \cite{ANALOGY:conf/icml/LiuWY17}	
    & 11.8 & 21.9 & 13.1 & 24.0 & 40.5		\\		
    \hline
    CrossE	 & \textbf{17.7} & \textbf{29.9} & \textbf{21.1}$^\dagger$ & \textbf{33.1}$^\dagger$ & \textbf{47.4}$^\dagger$	\\
    CrossE$_{S}$  & 6.40 & 11.0 & 6.7 & 11.7 & 19.8    \\
    \bottomrule

\end{tabular}
\caption{\small{Link prediction results on FB15k-237. }}
\label{LPdetail2}
\vspace{-5mm}
\end{table}

For WN18 (Table~\ref{link prediction results}), CrossE achieves \emph{Hit@10} results that are comparable to the best baselines. On the same dataset (Table~\ref{LPdetail}), CrossE achieves better results than majority of baselines on \emph{MRR}, \emph{Hit@1} and \emph{Hit@3}. With only $18$ relations, WN18 is a simpler dataset compared to FB15k. We can see in Table~\ref{LPdetail} that each method performs well on WN18 and much better than FB15k. For example, the scores of \emph{Hit@3} on WN18 with all methods are above $90\%$ while the best score on FB15k is just around $80\%$.
 
For FB15k (Table~\ref{link prediction results} and Table~\ref{LPdetail}) we see that CrossE achieves state-of-the-art results on majority evaluation metrics, including \emph{MRR}, \emph{Hit@3}, and \emph{Hit@10}. These results support that CrossE can encode the diverse relations in knowledge graph better, since FB15k is a complex linked dataset with $1,345$ relations and is more challenging than WN18. Compared to ANALOGY, which achieves best results on \emph{Hit@1}, CrossE performs better on \emph{Hit@3} and \emph{Hit@10}, 
two metrics that we think are better for datasets with diverse types of relations. In FB15k, there are $26.2\%$ 1-to-1 relations, $22.7\%$ 1-to-many, $28.3\%$ many-to-1 and $22.8\%$ many-to-many relations. 
Based on the Open World Assumption, the number of correct answer for link prediction on 1-to-many, many-to-1 and many-to-many relations might be more than $1$ even under the \emph{filter} setting.

For FB15k-237 (Table~\ref{LPdetail2}), CrossE achieves state-of-the-art results and significant improvements, compared with all baselines on all evaluation metrics. 
Compared to FB15k, FB15k-237 removes redundant triples causing inverse relation information leakage which can be easily captured by simpler approaches~\cite{Node-Leakf}. Thus without properly encoding the diverse semantics in the knowledge graph, a KGE can't achieve good performance on FB15k-237. The significant performance improvement achieved by CrossE indicates that it is more capable of capturing and utilizing the complex semantics in knowledge graph during prediction. 

\begin{figure*}[htbp] 
\centering 
\vspace{-8mm}
\subfigure[]{
	\label{recall-AvgSupport}
	\includegraphics[scale=0.555]{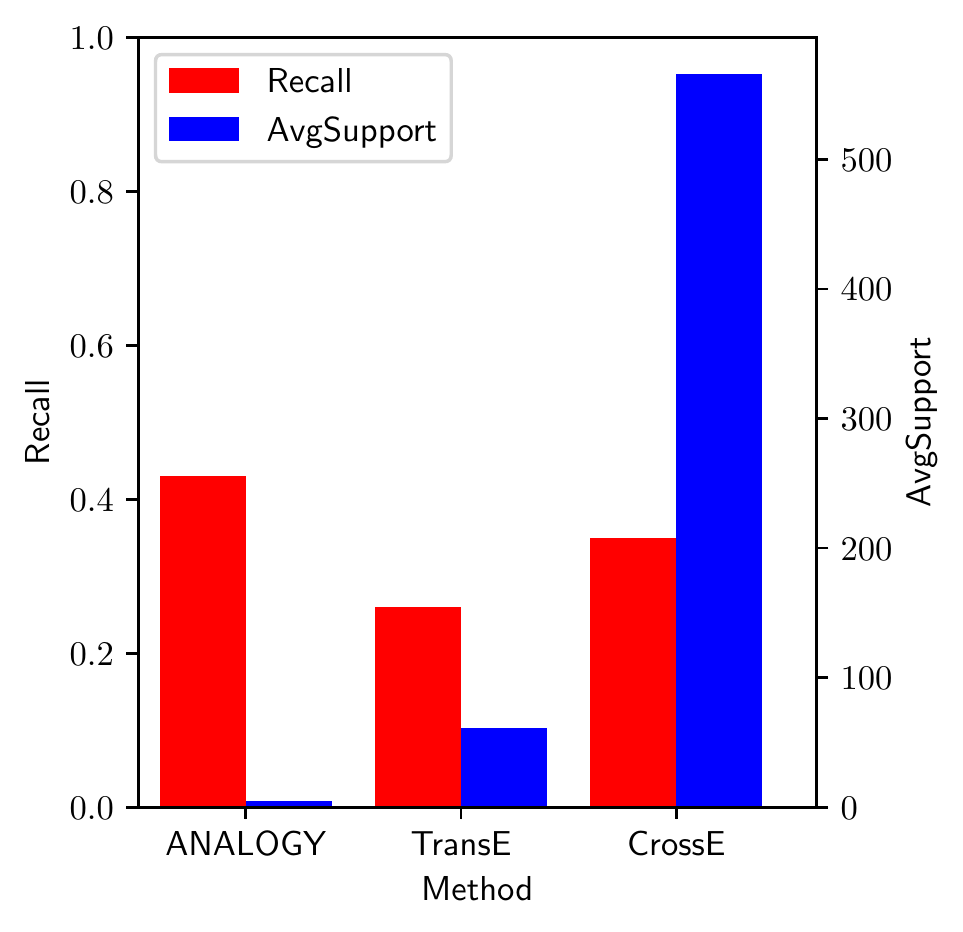}
	\vspace{-10mm}
}
\subfigure[]{
	\label{relation-variation}
	\includegraphics[scale=0.555]{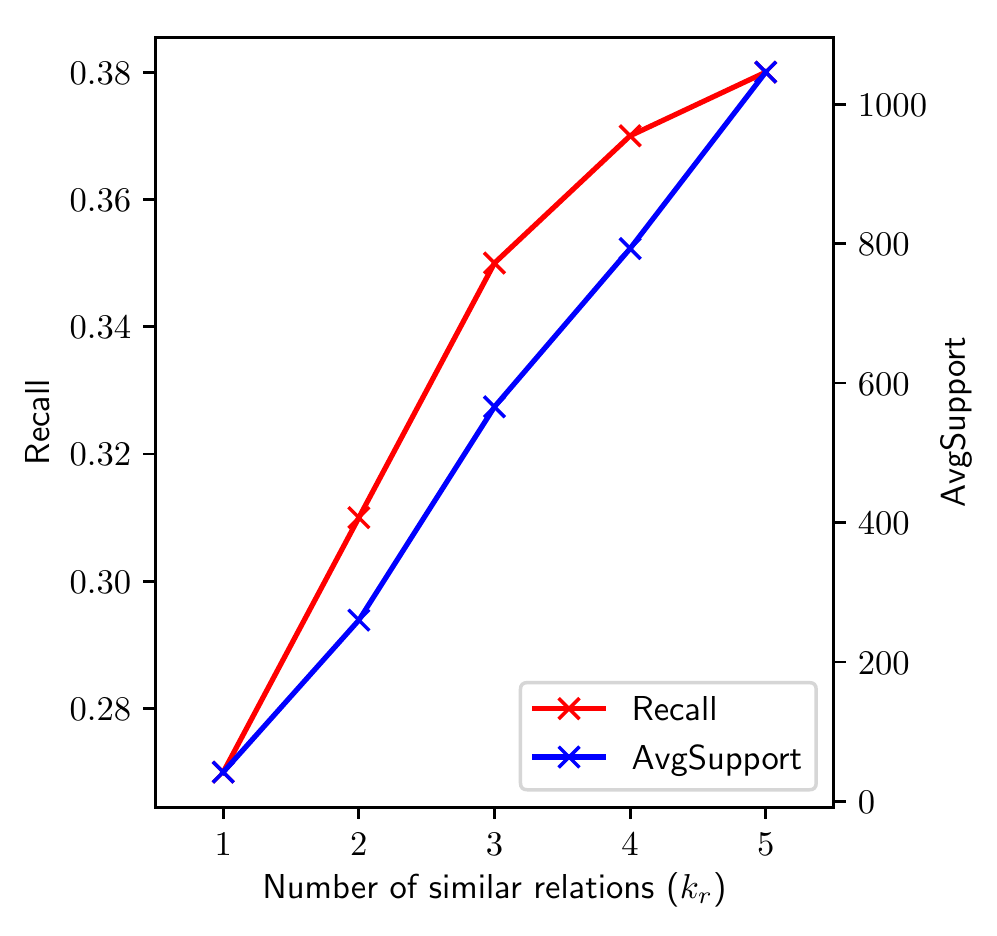}
	\vspace{-10mm}
}	
\subfigure[]{
	\label{support-vary}
	\includegraphics[scale=0.555]{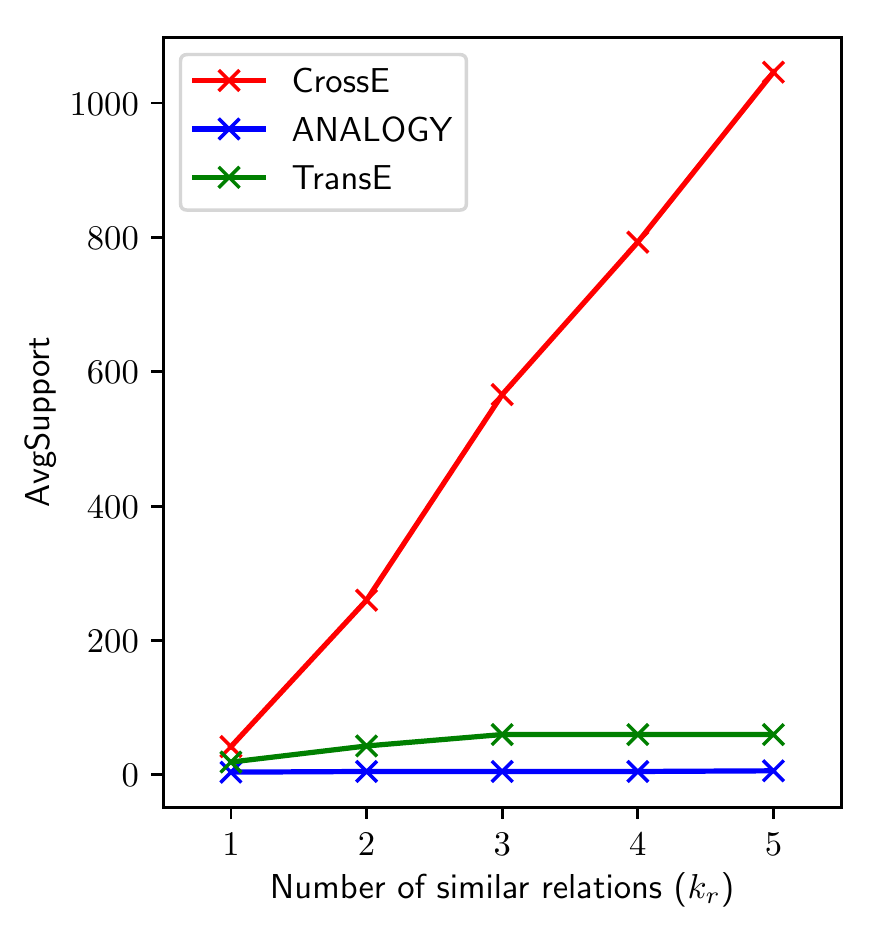}
\vspace{-10mm}
}
\vspace{-5mm}
\caption{
\small{
Evaluation results on generating explanations with different KGEs. \ref{recall-AvgSupport} shows \emph{Recall} and \emph{AvgSupport} with $k_e = 10$, $k_r = 3$. 
\ref{relation-variation} shows \emph{Recall} and \emph{AvgSupport} for CrossE with $k_r \in [1,5]$.
\ref{support-vary} shows \emph{AvgSupport} with $k_e = 10$ and $k_r \in [1,5]$.
}
}

\label{explanations} 
\vspace{-3mm}
\end{figure*} 

Compared to the simpler model CrossE$_{S}$, CrossE performs much better on all three datasets(Table~\ref{link prediction results}, Table~\ref{LPdetail}, and Table~\ref{LPdetail2}). As the only difference between them is whether there are \emph{crossover interactions} between entities and relations or not, the huge performance difference shows the importance of modeling \emph{crossover interactions}, which is common during inference on knowledge graphs with diverse topics.

To show which type of relations CrossE can encode better,
Table~\ref{relation type result} compares \emph{Hit@10 (filter)} results on FB15k after mapping different relation types, 1-to-1, 1-to-many, many-to-1 and many-to-many, represented as 1-1, 1-N, N-1 and N-N respectively.  
Separating rule of relation types follows \cite{TransE:conf/nips/BordesUGWY13}. 
\begin{table}[!htb]
\small
\centering
\begin{tabular}{c |c c c c}
    \toprule
 & \multicolumn{1}{c|}{1-1} & \multicolumn{1}{c|}{1-N} & \multicolumn{1}{c|}{N-1} & \multicolumn{1}{c}{N-N}   \\
 & \multicolumn{1}{c|}{(head/tail)} & \multicolumn{1}{c|}{(head/tail)} & \multicolumn{1}{c|}{(head/tail)} & \multicolumn{1}{c}{(head/tail)} \\
 \hline
\footnotesize{Unstructured\cite{Unstructured/SME:journals/ml/BordesGWB14}}      & 34.5/34.3   & 2.5/4.2     & 6.1/1.9    & 6.6/6.6     \\
SE\cite{SE:conf/aaai/BordesWCB11}              & 35.6/34.9   & 62.6/14.6   & 17.2/68.3    & 37.5/41.3     \\
\footnotesize{SME(linear)}\cite{Unstructured/SME:journals/ml/BordesGWB14}     & 35.1/32.7   & 53.7/14.9   & 19.0/61.6    & 40.3/43.3   \\
\footnotesize{SME(Bilinear)}\cite{Unstructured/SME:journals/ml/BordesGWB14}  & 30.9/28.2   & 69.6/13.1   & 19.9/76.0   & 38.6/41.8    \\
 TransE\cite{TransE:conf/nips/BordesUGWY13}         & 43.7/43.7   & 65.7/19.7   & 18.2/66.7   & 47.2/50.0  \\
  TransH\cite{TransH:conf/aaai/WangZFC14}   & 66.8/65.5   & 87.6/39.8   & 28.7/83.3   & 64.5/67.2   \\
 TransD\cite{TransD:conf/acl/JiHXL015}    & 86.1/85.4   & 95.5/50.6   & 39.8/\textbf{94.4}   & 78.5/81.2 \\
 TransR\cite{TransR:conf/aaai/LinLSLZ15}     & 78.8/79.2   & 89.2/37.4   & 34.1/90.4   & 69.2/72.1     \\
 CTransR\cite{TransR:conf/aaai/LinLSLZ15}    & 81.5/80.8   & 89.0/38.6    & 34.7/90.1   & 71.2/73.8   \\
 \hline
  CrossE    & \underline{\textbf{88.2}}$^\dagger$/\underline{\textbf{87.7}}$^\dagger$   &  \underline{\textbf{95.7}}$^\dagger$/\underline{\textbf{75.1}}$^\dagger$  &   \underline{\textbf{64.2}}$^\dagger$/\underline{92.3}	  & 	\underline{\textbf{88.1}}$^\dagger$/\underline{\textbf{90.8}}$^\dagger$ \\
   CrossE$_{S}$      & 78.6/81.6 &85.1/54.2  &45.3 /85.8 & 71.7/76.7 \\
    \bottomrule
 \end{tabular}
  \caption{\small{Hit@10 on FB15k by mapping to different relation types.}}
 \label{relation type result}
 \vspace{-4mm}
\end{table}

From Table~\ref{relation type result}, we see that CrossE significantly outperforms all other embedding methods except in the tail prediction for N-1 relations. 
On more difficult tasks with more correct answers including head prediction for N-1 relations, tail prediction for 1-N relations and both head and tail prediction for N-N relations, CrossE achieves significant improvement, with $11.7\%$ in average. 
As a conclusion, CrossE performs more stably than other methods on different types of relations.

%!TEX root = crosse_wsdm.tex
\vspace{-3mm}
\subsection{Evaluation \uppercase\expandafter{\romannumeral2}: Generating Explanations}
\label{sec:explain_eval}

\subsubsection{Evaluation}
Aggregating all path explanations and their supports from embedding results,
we evaluate the capability of a KGE to generate explanations from two perspectives:
(1) the fraction of triples (out of all test triples) that KGE can give explanations for (\emph{Recall}),
(2) the average support among the triples for which it can find explanations (\emph{AvgSupport}). 
We argue that higher the \emph{AvgSupport} for a triple, the more reliable the explanation will be. 

Generally, a KGE that can generate better explanations will achieve higher \emph{Recall} and \emph{AvgSupport} when selecting the same number of similar entities ($k_e$) and relations ($k_r$). 
\vspace{-2mm}
\subsubsection{Experimental Details}
For the explanation experiment, we use the FB15k dataset.
We choose two KGEs as baselines. One is the popular translation-based embedding method TransE, and the other is the linear-mapping based embedding method ANALOGY which achieves one of the best results on link prediction tasks.

The embeddings of CrossE used for searching explanations are those used in the link prediction experiment.
We re-implement TransE and the implementation of ANALOGY is from~\cite{ANALOGY:conf/icml/LiuWY17}.  
During the similar entity and relation selection, embeddings for head entities and relations when they involve specific triples are used, which are \crossembed in CrossE and general embeddings in TransE and ANALOGY.

\vspace{-2mm}
\subsubsection{Explanation Results}
The results are summarized in Figure~\ref{explanations}. 
In Figure~\ref{recall-AvgSupport}, we see that when selecting ten similar entities and three similar relations, the \emph{recall} of three methods varies from $0.26$ to $0.43$ and the \emph{AvgSupport} varies from $5$ to $566$. ANALOGY achieves the best results on \emph{Recall} while it can give only a few examples for each explanation, quantified by its \emph{AvgSupport} performance. TransE achieves the lowest \emph{recall} but its \emph{AvgSupport} for each triple is about 10 times more than ANALOGY.
CrossE achieves the second best result on \emph{recall} and has about $100$ times more \emph{AvgSupport} than ANALOGY. From the perspective of giving reliable explanations for predicted triples, CrossE outperforms the baselines.

The explanation and similar structures search are based on the similar entity and relation selection results. 
CrossE generates multiple crossover embeddings for entities and relations. 
Based on the \crossembed, the similar items for each entity or relation are different when it involves different triples. 
This makes CrossE more efficient in selecting similar items, and will result in giving explanations better.
Based on general embeddings, similar items selection results will always be the same, regardless of the triple-specific context.
In our opinion, this is why the \emph{AvgSupport} of CrossE is much higher than of TransE and ANALOGY.

In Figure \ref{relation-variation}, \emph{recall} increases slightly when selecting more similar relations while \emph{AvgSupport} increases a lot. It is the same when increasing the number of selected similar entities. Figure~\ref{support-vary} shows that \emph{AvgSupport} for CrossE increases much faster than that for TransE and ANALOGY. This also demonstrates the effectiveness of CrossE at selecting similar entities and relations. 

\begin{figure}[htbp] 
\vspace{-2mm}
\centering 
\includegraphics[scale=0.6]{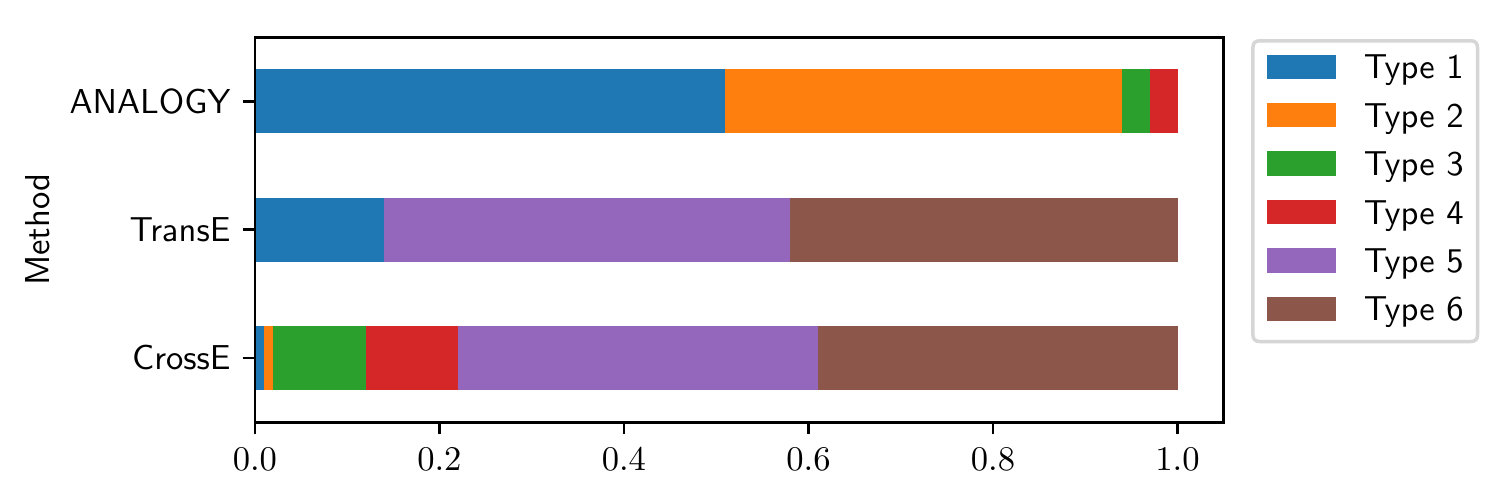}
\vspace{-8mm}
\caption{\small{Share of \emph{AvgSupport} for six similar structure types.}} 
\label{type-percent}
\vspace{-3mm}
\end{figure}

To figure out what types of paths and similar structures are easier to find for compared models, in Figure~\ref{type-percent}, for each model, we show the share of \emph{AvgSupport} for each type of similar structures. 

We can see that among all types of similar structures, type 5 is the one that both TransE and CrossE have the most \emph{AvgSupport} for. From our point of view, type 5 is the most natural path, where two relations along the path are in the same direction with the relation between $h$ and $t$. 
Thus type 5 is more likely to be constructed when building a knowledge graph.
Although the shares of type 1 and 2 are high for ANALOGY, the \emph{AvgSupport} values are very low (Figure~\ref{support-vary}).  

In summary, we can state that the design of KGE models and their vector-space assumptions will affect the type of path-explanations they can provide for.

From these two evaluation tasks, we can conclude that the capability of KGE on link prediction and explanations are not directly related. 
A method that performs well on link prediction may not necessarily be good at giving explanations. Thus the balance between prediction accuracy and giving explanations is important.

%!TEX root = crosse_wsdm.tex
\vspace{-2mm}
\section{Conclusion}
\label{sec:conclusion}
In this paper, we described a new knowledge graph embedding named CrossE. CrossE successfully captures crossover interactions between entities and relations when modeling knowledge graphs and achieves state-of-the-art results on link prediction task with complex linked datasets. We believe that improving the reliability of embedding method is as important as achieving high-accuracy prediction. This work is a first step for explaining the prediction results. There are still much work to do with explanations, such as how to enable KGEs to give explanations for all predicted triples. In our future work, we will focus on improving the capability of KGEs in predict missing triples and also giving more reliable explanations.% for the predicted results.

\begin{acks}
\vspace{-1mm}
This work is funded by NSFC 61673338/61473260, and supported by Alibaba-Zhejiang University Joint Institute of Frontier Technologies and SNF Sino Swiss Science and Technology Cooperation Programme program under contract RiC 01-032014.
\end{acks}

\bibliographystyle{ACM-Reference-Format}
\bibliography{embedding}


\begin{thebibliography}{44}

%%% the Web address.

\ifx \showCODEN    \undefined \def \showCODEN     #1{\unskip}     \fi
\ifx \showDOI      \undefined \def \showDOI       #1{#1}\fi
\ifx \showISBNx    \undefined \def \showISBNx     #1{\unskip}     \fi
\ifx \showISBNxiii \undefined \def \showISBNxiii  #1{\unskip}     \fi
\ifx \showISSN     \undefined \def \showISSN      #1{\unskip}     \fi
\ifx \showLCCN     \undefined \def \showLCCN      #1{\unskip}     \fi
\ifx \shownote     \undefined \def \shownote      #1{#1}          \fi
\ifx \showarticletitle \undefined \def \showarticletitle #1{#1}   \fi
\ifx \showURL      \undefined \def \showURL       {\relax}        \fi
% invisible to TeX
\providecommand\bibfield[2]{#2}
\providecommand\bibinfo[2]{#2}
\providecommand\natexlab[1]{#1}
\providecommand\showeprint[2][]{arXiv:#2}

\bibitem[\protect\citeauthoryear{Bollacker, Evans, Paritosh, Sturge, and
  Taylor}{Bollacker et~al\mbox{.}}{2008}]%
        {Freebase:conf/sigmod/BollackerEPST08}
\bibfield{author}{\bibinfo{person}{Kurt~D. Bollacker}, \bibinfo{person}{Colin
  Evans}, \bibinfo{person}{Praveen Paritosh}, \bibinfo{person}{Tim Sturge},
  {and} \bibinfo{person}{Jamie Taylor}.} \bibinfo{year}{2008}\natexlab{}.
\newblock \showarticletitle{Freebase: a collaboratively created graph database
  for structuring human knowledge}.
\newblock \bibinfo{journal}{\emph{Proceedings of {SIGMOD}}}
  (\bibinfo{year}{2008}), \bibinfo{pages}{1247--1250}.
\newblock


\bibitem[\protect\citeauthoryear{Bordes, Glorot, Weston, and Bengio}{Bordes
  et~al\mbox{.}}{2014a}]%
        {Unstructed/SME:journals/ml/BordesGWB14}
\bibfield{author}{\bibinfo{person}{Antoine Bordes}, \bibinfo{person}{Xavier
  Glorot}, \bibinfo{person}{Jason Weston}, {and} \bibinfo{person}{Yoshua
  Bengio}.} \bibinfo{year}{2014}\natexlab{a}.
\newblock \showarticletitle{A semantic matching energy function for learning
  with multi-relational data - Application to word-sense disambiguation}.
\newblock \bibinfo{journal}{\emph{Machine Learning}} \bibinfo{volume}{94},
  \bibinfo{number}{2} (\bibinfo{year}{2014}), \bibinfo{pages}{233--259}.
\newblock


\bibitem[\protect\citeauthoryear{Bordes, Glorot, Weston, and Bengio}{Bordes
  et~al\mbox{.}}{2014b}]%
        {Unstructured/SME:journals/ml/BordesGWB14}
\bibfield{author}{\bibinfo{person}{Antoine Bordes}, \bibinfo{person}{Xavier
  Glorot}, \bibinfo{person}{Jason Weston}, {and} \bibinfo{person}{Yoshua
  Bengio}.} \bibinfo{year}{2014}\natexlab{b}.
\newblock \showarticletitle{A semantic matching energy function for learning
  with multi-relational data - Application to word-sense disambiguation}.
\newblock \bibinfo{journal}{\emph{Machine Learning}} \bibinfo{volume}{94},
  \bibinfo{number}{2} (\bibinfo{year}{2014}), \bibinfo{pages}{233--259}.
\newblock


\bibitem[\protect\citeauthoryear{Bordes, Usunier, Garc{\'{\i}}a{-}Dur{\'{a}}n,
  Weston, and Yakhnenko}{Bordes et~al\mbox{.}}{2013}]%
        {TransE:conf/nips/BordesUGWY13}
\bibfield{author}{\bibinfo{person}{Antoine Bordes}, \bibinfo{person}{Nicolas
  Usunier}, \bibinfo{person}{Alberto Garc{\'{\i}}a{-}Dur{\'{a}}n},
  \bibinfo{person}{Jason Weston}, {and} \bibinfo{person}{Oksana Yakhnenko}.}
  \bibinfo{year}{2013}\natexlab{}.
\newblock \showarticletitle{Translating Embeddings for Modeling
  Multi-relational Data}.
\newblock \bibinfo{journal}{\emph{Proceedings of {NIPS}}}
  (\bibinfo{year}{2013}), \bibinfo{pages}{2787--2795}.
\newblock


\bibitem[\protect\citeauthoryear{Bordes, Weston, Collobert, and Bengio}{Bordes
  et~al\mbox{.}}{2011}]%
        {SE:conf/aaai/BordesWCB11}
\bibfield{author}{\bibinfo{person}{Antoine Bordes}, \bibinfo{person}{Jason
  Weston}, \bibinfo{person}{Ronan Collobert}, {and} \bibinfo{person}{Yoshua
  Bengio}.} \bibinfo{year}{2011}\natexlab{}.
\newblock \showarticletitle{Learning Structured Embeddings of Knowledge Bases}.
\newblock \bibinfo{journal}{\emph{Proceedings of {AAAI}}}
  (\bibinfo{year}{2011}).
\newblock


\bibitem[\protect\citeauthoryear{Gal{\'{a}}rraga, Teflioudi, Hose, and
  Suchanek}{Gal{\'{a}}rraga et~al\mbox{.}}{2015}]%
        {AMIE+:journals/vldb/GalarragaTHS15}
\bibfield{author}{\bibinfo{person}{Luis Gal{\'{a}}rraga},
  \bibinfo{person}{Christina Teflioudi}, \bibinfo{person}{Katja Hose}, {and}
  \bibinfo{person}{Fabian~M. Suchanek}.} \bibinfo{year}{2015}\natexlab{}.
\newblock \showarticletitle{Fast rule mining in ontological knowledge bases
  with {AMIE+}}.
\newblock \bibinfo{journal}{\emph{{VLDB} J.}} \bibinfo{volume}{24},
  \bibinfo{number}{6} (\bibinfo{year}{2015}), \bibinfo{pages}{707--730}.
\newblock


\bibitem[\protect\citeauthoryear{Garc{\'{\i}}a{-}Dur{\'{a}}n, Bordes, and
  Usunier}{Garc{\'{\i}}a{-}Dur{\'{a}}n et~al\mbox{.}}{2015}]%
        {RTransE:conf/emnlp/Garcia-DuranBU15}
\bibfield{author}{\bibinfo{person}{Alberto Garc{\'{\i}}a{-}Dur{\'{a}}n},
  \bibinfo{person}{Antoine Bordes}, {and} \bibinfo{person}{Nicolas Usunier}.}
  \bibinfo{year}{2015}\natexlab{}.
\newblock \showarticletitle{Composing Relationships with Translations}.
\newblock \bibinfo{journal}{\emph{Proceedings of {EMNLP}}}
  (\bibinfo{year}{2015}), \bibinfo{pages}{286--290}.
\newblock


\bibitem[\protect\citeauthoryear{Glorot and Bengio}{Glorot and Bengio}{2010}]%
        {Init:journals/jmlr/GlorotB10}
\bibfield{author}{\bibinfo{person}{Xavier Glorot} {and} \bibinfo{person}{Yoshua
  Bengio}.} \bibinfo{year}{2010}\natexlab{}.
\newblock \showarticletitle{Understanding the difficulty of training deep
  feedforward neural networks}.
\newblock \bibinfo{journal}{\emph{Proceedings of {AISTATS}}}
  (\bibinfo{year}{2010}), \bibinfo{pages}{249--256}.
\newblock


\bibitem[\protect\citeauthoryear{He, Liu, Ji, and Zhao}{He
  et~al\mbox{.}}{2015}]%
        {KG2E:conf/cikm/HeLJ015}
\bibfield{author}{\bibinfo{person}{Shizhu He}, \bibinfo{person}{Kang Liu},
  \bibinfo{person}{Guoliang Ji}, {and} \bibinfo{person}{Jun Zhao}.}
  \bibinfo{year}{2015}\natexlab{}.
\newblock \showarticletitle{Learning to Represent Knowledge Graphs with
  Gaussian Embedding}.
\newblock \bibinfo{journal}{\emph{Proceedings of {CIKM}}}
  (\bibinfo{year}{2015}), \bibinfo{pages}{623--632}.
\newblock


\bibitem[\protect\citeauthoryear{Heckel, Vlachos, Parnell, and
  D{\"{u}}nner}{Heckel et~al\mbox{.}}{2017}]%
        {OCuLaR:conf/icde/HeckelVPD17}
\bibfield{author}{\bibinfo{person}{Reinhard Heckel}, \bibinfo{person}{Michail
  Vlachos}, \bibinfo{person}{Thomas~P. Parnell}, {and}
  \bibinfo{person}{Celestine D{\"{u}}nner}.} \bibinfo{year}{2017}\natexlab{}.
\newblock \showarticletitle{Scalable and Interpretable Product Recommendations
  via Overlapping Co-Clustering}.
\newblock \bibinfo{journal}{\emph{Procedding of {ICDE}}}
  (\bibinfo{year}{2017}), \bibinfo{pages}{1033--1044}.
\newblock


\bibitem[\protect\citeauthoryear{Hinton, McClelland, and Rumelhart}{Hinton
  et~al\mbox{.}}{1986}]%
        {RepresentationLearning/Hinton/1986}
\bibfield{author}{\bibinfo{person}{G.~E. Hinton}, \bibinfo{person}{J.~L.
  McClelland}, {and} \bibinfo{person}{D.~E. Rumelhart}.}
  \bibinfo{year}{1986}\natexlab{}.
\newblock \showarticletitle{Distributed Representations}.
\newblock \bibinfo{journal}{\emph{Parallel distributed processing: explorations
  in the microstructure of cognition}} (\bibinfo{year}{1986}).
\newblock


\bibitem[\protect\citeauthoryear{Jenatton, Roux, Bordes, and
  Obozinski}{Jenatton et~al\mbox{.}}{2012}]%
        {LFM:conf/nips/JenattonRBO12}
\bibfield{author}{\bibinfo{person}{Rodolphe Jenatton},
  \bibinfo{person}{Nicolas~Le Roux}, \bibinfo{person}{Antoine Bordes}, {and}
  \bibinfo{person}{Guillaume Obozinski}.} \bibinfo{year}{2012}\natexlab{}.
\newblock \showarticletitle{A latent factor model for highly multi-relational
  data}.
\newblock \bibinfo{journal}{\emph{Proceddings of {NIPS}}}
  (\bibinfo{year}{2012}), \bibinfo{pages}{3176--3184}.
\newblock


\bibitem[\protect\citeauthoryear{Ji, He, Xu, Liu, and Zhao}{Ji
  et~al\mbox{.}}{2015}]%
        {TransD:conf/acl/JiHXL015}
\bibfield{author}{\bibinfo{person}{Guoliang Ji}, \bibinfo{person}{Shizhu He},
  \bibinfo{person}{Liheng Xu}, \bibinfo{person}{Kang Liu}, {and}
  \bibinfo{person}{Jun Zhao}.} \bibinfo{year}{2015}\natexlab{}.
\newblock \showarticletitle{Knowledge Graph Embedding via Dynamic Mapping
  Matrix}.
\newblock \bibinfo{journal}{\emph{Proceedings of {ACL}}}
  (\bibinfo{year}{2015}), \bibinfo{pages}{687--696}.
\newblock


\bibitem[\protect\citeauthoryear{Ji, Liu, He, and Zhao}{Ji
  et~al\mbox{.}}{2016}]%
        {TranSparse:conf/aaai/JiLH016}
\bibfield{author}{\bibinfo{person}{Guoliang Ji}, \bibinfo{person}{Kang Liu},
  \bibinfo{person}{Shizhu He}, {and} \bibinfo{person}{Jun Zhao}.}
  \bibinfo{year}{2016}\natexlab{}.
\newblock \showarticletitle{Knowledge Graph Completion with Adaptive Sparse
  Transfer Matrix}.
\newblock \bibinfo{journal}{\emph{Proceedings of {AAAI}}}
  (\bibinfo{year}{2016}), \bibinfo{pages}{985--991}.
\newblock


\bibitem[\protect\citeauthoryear{Kingma and Ba}{Kingma and Ba}{2014}]%
        {Adam:journals/corr/KingmaB14}
\bibfield{author}{\bibinfo{person}{Diederik~P. Kingma} {and}
  \bibinfo{person}{Jimmy Ba}.} \bibinfo{year}{2014}\natexlab{}.
\newblock \showarticletitle{Adam: {A} Method for Stochastic Optimization}.
\newblock \bibinfo{journal}{\emph{CoRR}}  \bibinfo{volume}{abs/1412.6980}
  (\bibinfo{year}{2014}).
\newblock


\bibitem[\protect\citeauthoryear{Krompa{\ss}, Baier, and Tresp}{Krompa{\ss}
  et~al\mbox{.}}{2015}]%
        {TypeConstrain:conf/semweb/KrompassBT15}
\bibfield{author}{\bibinfo{person}{Denis Krompa{\ss}}, \bibinfo{person}{Stephan
  Baier}, {and} \bibinfo{person}{Volker Tresp}.}
  \bibinfo{year}{2015}\natexlab{}.
\newblock \showarticletitle{Type-Constrained Representation Learning in
  Knowledge Graphs}.
\newblock \bibinfo{journal}{\emph{Proceddings of {ISWC}}}
  (\bibinfo{year}{2015}), \bibinfo{pages}{640--655}.
\newblock


\bibitem[\protect\citeauthoryear{Lao, Mitchell, and Cohen}{Lao
  et~al\mbox{.}}{2011}]%
        {PRA:conf/emnlp/LaoMC11}
\bibfield{author}{\bibinfo{person}{Ni Lao}, \bibinfo{person}{Tom~M. Mitchell},
  {and} \bibinfo{person}{William~W. Cohen}.} \bibinfo{year}{2011}\natexlab{}.
\newblock \showarticletitle{Random Walk Inference and Learning in {A} Large
  Scale Knowledge Base}. In \bibinfo{booktitle}{\emph{{EMNLP}}}.
  \bibinfo{publisher}{{ACL}}, \bibinfo{pages}{529--539}.
\newblock


\bibitem[\protect\citeauthoryear{Lin, Liu, Luan, Sun, Rao, and Liu}{Lin
  et~al\mbox{.}}{2015a}]%
        {PTransE:conf/emnlp/LinLLSRL15}
\bibfield{author}{\bibinfo{person}{Yankai Lin}, \bibinfo{person}{Zhiyuan Liu},
  \bibinfo{person}{Huan{-}Bo Luan}, \bibinfo{person}{Maosong Sun},
  \bibinfo{person}{Siwei Rao}, {and} \bibinfo{person}{Song Liu}.}
  \bibinfo{year}{2015}\natexlab{a}.
\newblock \showarticletitle{Modeling Relation Paths for Representation Learning
  of Knowledge Bases}.
\newblock \bibinfo{journal}{\emph{Proceedings of {EMNLP}}}
  (\bibinfo{year}{2015}), \bibinfo{pages}{705--714}.
\newblock


\bibitem[\protect\citeauthoryear{Lin, Liu, Sun, Liu, and Zhu}{Lin
  et~al\mbox{.}}{2015b}]%
        {TransR:conf/aaai/LinLSLZ15}
\bibfield{author}{\bibinfo{person}{Yankai Lin}, \bibinfo{person}{Zhiyuan Liu},
  \bibinfo{person}{Maosong Sun}, \bibinfo{person}{Yang Liu}, {and}
  \bibinfo{person}{Xuan Zhu}.} \bibinfo{year}{2015}\natexlab{b}.
\newblock \showarticletitle{Learning Entity and Relation Embeddings for
  Knowledge Graph Completion}.
\newblock \bibinfo{journal}{\emph{Proceedings of {AAAI}}}
  (\bibinfo{year}{2015}), \bibinfo{pages}{2181--2187}.
\newblock


\bibitem[\protect\citeauthoryear{Liu, Wu, and Yang}{Liu et~al\mbox{.}}{2017}]%
        {ANALOGY:conf/icml/LiuWY17}
\bibfield{author}{\bibinfo{person}{Hanxiao Liu}, \bibinfo{person}{Yuexin Wu},
  {and} \bibinfo{person}{Yiming Yang}.} \bibinfo{year}{2017}\natexlab{}.
\newblock \showarticletitle{Analogical Inference for Multi-relational
  Embeddings}. In \bibinfo{booktitle}{\emph{Proceedings of the 34th
  International Conference on Machine Learning, {ICML} 2017, Sydney, NSW,
  Australia, 6-11 August 2017}}. \bibinfo{pages}{2168--2178}.
\newblock


\bibitem[\protect\citeauthoryear{Miller}{Miller}{1995}]%
        {WordNet:journals/cacm/Miller95}
\bibfield{author}{\bibinfo{person}{George~A. Miller}.}
  \bibinfo{year}{1995}\natexlab{}.
\newblock \showarticletitle{WordNet: {A} Lexical Database for English}.
\newblock \bibinfo{journal}{\emph{Commun. {ACM}}} \bibinfo{volume}{38},
  \bibinfo{number}{11} (\bibinfo{year}{1995}), \bibinfo{pages}{39--41}.
\newblock


\bibitem[\protect\citeauthoryear{Neelakantan, Roth, and McCallum}{Neelakantan
  et~al\mbox{.}}{2015}]%
        {CVSM:conf/acl/NeelakantanRM15}
\bibfield{author}{\bibinfo{person}{Arvind Neelakantan},
  \bibinfo{person}{Benjamin Roth}, {and} \bibinfo{person}{Andrew McCallum}.}
  \bibinfo{year}{2015}\natexlab{}.
\newblock \showarticletitle{Compositional Vector Space Models for Knowledge
  Base Completion}.
\newblock \bibinfo{journal}{\emph{Proceedings of {ACL}}}
  (\bibinfo{year}{2015}), \bibinfo{pages}{156--166}.
\newblock


\bibitem[\protect\citeauthoryear{Nguyen, Sirts, Qu, and Johnson}{Nguyen
  et~al\mbox{.}}{2016}]%
        {STransE:conf/naacl/NguyenSQJ16}
\bibfield{author}{\bibinfo{person}{Dat~Quoc Nguyen}, \bibinfo{person}{Kairit
  Sirts}, \bibinfo{person}{Lizhen Qu}, {and} \bibinfo{person}{Mark Johnson}.}
  \bibinfo{year}{2016}\natexlab{}.
\newblock \showarticletitle{STransE: a novel embedding model of entities and
  relationships in knowledge bases}. In
  \bibinfo{booktitle}{\emph{{HLT-NAACL}}}. \bibinfo{publisher}{The Association
  for Computational Linguistics}, \bibinfo{pages}{460--466}.
\newblock


\bibitem[\protect\citeauthoryear{Nickel, Murphy, Tresp, and Gabrilovich}{Nickel
  et~al\mbox{.}}{2016a}]%
        {MLKG:journals/pieee/Nickel0TG16}
\bibfield{author}{\bibinfo{person}{Maximilian Nickel}, \bibinfo{person}{Kevin
  Murphy}, \bibinfo{person}{Volker Tresp}, {and} \bibinfo{person}{Evgeniy
  Gabrilovich}.} \bibinfo{year}{2016}\natexlab{a}.
\newblock \showarticletitle{A Review of Relational Machine Learning for
  Knowledge Graphs}.
\newblock \bibinfo{journal}{\emph{Proc. IEEE}} \bibinfo{volume}{104},
  \bibinfo{number}{1} (\bibinfo{year}{2016}), \bibinfo{pages}{11--33}.
\newblock


\bibitem[\protect\citeauthoryear{Nickel, Rosasco, and Poggio}{Nickel
  et~al\mbox{.}}{2016b}]%
        {HolE:conf/aaai/NickelRP16}
\bibfield{author}{\bibinfo{person}{Maximilian Nickel}, \bibinfo{person}{Lorenzo
  Rosasco}, {and} \bibinfo{person}{Tomaso~A. Poggio}.}
  \bibinfo{year}{2016}\natexlab{b}.
\newblock \showarticletitle{Holographic Embeddings of Knowledge Graphs}.
\newblock \bibinfo{journal}{\emph{Proceedings of {AAAI}}}
  (\bibinfo{year}{2016}), \bibinfo{pages}{1955--1961}.
\newblock


\bibitem[\protect\citeauthoryear{Nickel, Tresp, and Kriegel}{Nickel
  et~al\mbox{.}}{2011}]%
        {RESCAL:conf/icml/NickelTK11}
\bibfield{author}{\bibinfo{person}{Maximilian Nickel}, \bibinfo{person}{Volker
  Tresp}, {and} \bibinfo{person}{Hans{-}Peter Kriegel}.}
  \bibinfo{year}{2011}\natexlab{}.
\newblock \showarticletitle{A Three-Way Model for Collective Learning on
  Multi-Relational Data}.
\newblock \bibinfo{journal}{\emph{Proceedings of {ICML}}}
  (\bibinfo{year}{2011}), \bibinfo{pages}{809--816}.
\newblock


\bibitem[\protect\citeauthoryear{Ristoski and Paulheim}{Ristoski and
  Paulheim}{2016}]%
        {RDF2Vec:conf/semweb/RistoskiP16}
\bibfield{author}{\bibinfo{person}{Petar Ristoski} {and} \bibinfo{person}{Heiko
  Paulheim}.} \bibinfo{year}{2016}\natexlab{}.
\newblock \showarticletitle{RDF2Vec: {RDF} Graph Embeddings for Data Mining}.
\newblock \bibinfo{journal}{\emph{Procedings of {ISWC}}}
  (\bibinfo{year}{2016}), \bibinfo{pages}{498--514}.
\newblock


\bibitem[\protect\citeauthoryear{Schlichtkrull, Kipf, Bloem, van~den Berg,
  Titov, and Welling}{Schlichtkrull et~al\mbox{.}}{2018}]%
        {R-GCN:conf/eswc}
\bibfield{author}{\bibinfo{person}{Michael Schlichtkrull},
  \bibinfo{person}{Thomas~N. Kipf}, \bibinfo{person}{Peter Bloem},
  \bibinfo{person}{Rianne van~den Berg}, \bibinfo{person}{Ivan Titov}, {and}
  \bibinfo{person}{Max Welling}.} \bibinfo{year}{2018}\natexlab{}.
\newblock \showarticletitle{Modeling Relational Data with Graph Convolutional
  Networks}.
\newblock \bibinfo{journal}{\emph{Proceddings of {ESWC}}}
  (\bibinfo{year}{2018}).
\newblock


\bibitem[\protect\citeauthoryear{Shi and Weninger}{Shi and Weninger}{2017}]%
        {ProjE:conf/aaai/ShiW17}
\bibfield{author}{\bibinfo{person}{Baoxu Shi} {and} \bibinfo{person}{Tim
  Weninger}.} \bibinfo{year}{2017}\natexlab{}.
\newblock \showarticletitle{ProjE: Embedding Projection for Knowledge Graph
  Completion}.
\newblock \bibinfo{journal}{\emph{Proceedings of {AAAI}}}
  (\bibinfo{year}{2017}), \bibinfo{pages}{1236--1242}.
\newblock


\bibitem[\protect\citeauthoryear{Socher, Chen, Manning, and Ng}{Socher
  et~al\mbox{.}}{2013}]%
        {NTN:conf/nips/SocherCMN13}
\bibfield{author}{\bibinfo{person}{Richard Socher}, \bibinfo{person}{Danqi
  Chen}, \bibinfo{person}{Christopher~D. Manning}, {and}
  \bibinfo{person}{Andrew~Y. Ng}.} \bibinfo{year}{2013}\natexlab{}.
\newblock \showarticletitle{Reasoning With Neural Tensor Networks for Knowledge
  Base Completion}.
\newblock \bibinfo{journal}{\emph{Prodeddings of {NIPS}}}
  (\bibinfo{year}{2013}), \bibinfo{pages}{926--934}.
\newblock


\bibitem[\protect\citeauthoryear{Srivastava, Hinton, Krizhevsky, Sutskever, and
  Salakhutdinov}{Srivastava et~al\mbox{.}}{2014}]%
        {dropout:journal/JMLR/2014}
\bibfield{author}{\bibinfo{person}{Nitish Srivastava},
  \bibinfo{person}{Geoffrey Hinton}, \bibinfo{person}{Alex Krizhevsky},
  \bibinfo{person}{Ilya Sutskever}, {and} \bibinfo{person}{Ruslan
  Salakhutdinov}.} \bibinfo{year}{2014}\natexlab{}.
\newblock \showarticletitle{Dropout: A Simple Way to Prevent Neural Networks
  from Overfitting}.
\newblock \bibinfo{journal}{\emph{Journal of Machine Learning Research
  15(Jun)}} (\bibinfo{year}{2014}), \bibinfo{pages}{1929--1958}.
\newblock


\bibitem[\protect\citeauthoryear{Suchanek, Kasneci, and Weikum}{Suchanek
  et~al\mbox{.}}{2007}]%
        {YAGO:conf/www/SuchanekKW07}
\bibfield{author}{\bibinfo{person}{Fabian~M. Suchanek},
  \bibinfo{person}{Gjergji Kasneci}, {and} \bibinfo{person}{Gerhard Weikum}.}
  \bibinfo{year}{2007}\natexlab{}.
\newblock \showarticletitle{Yago: a core of semantic knowledge}.
\newblock \bibinfo{journal}{\emph{Proceedings of {WWW}}}
  (\bibinfo{year}{2007}), \bibinfo{pages}{697--706}.
\newblock


\bibitem[\protect\citeauthoryear{Szumlanski and Gomez}{Szumlanski and
  Gomez}{2010}]%
        {WebSearch:conf/cikm/SzumlanskiG10}
\bibfield{author}{\bibinfo{person}{Sean~R. Szumlanski} {and}
  \bibinfo{person}{Fernando Gomez}.} \bibinfo{year}{2010}\natexlab{}.
\newblock \showarticletitle{Automatically acquiring a semantic network of
  related concepts}.
\newblock  (\bibinfo{year}{2010}), \bibinfo{pages}{19--28}.
\newblock


\bibitem[\protect\citeauthoryear{Toutanova and Chen}{Toutanova and
  Chen}{2015}]%
        {Node-Leakf}
\bibfield{author}{\bibinfo{person}{Kristina Toutanova} {and}
  \bibinfo{person}{Danqi Chen}.} \bibinfo{year}{2015}\natexlab{}.
\newblock \showarticletitle{Observed versus latent features for knowledge base
  and text inference}. In \bibinfo{booktitle}{\emph{Proceedings of the 3rd
  Workshop on Continuous Vector Space Models and their Compositionality}}.
  \bibinfo{pages}{57--66}.
\newblock


\bibitem[\protect\citeauthoryear{Trouillon, Welbl, Riedel, Gaussier, and
  Bouchard}{Trouillon et~al\mbox{.}}{2016}]%
        {ComplEx:conf/icml/TrouillonWRGB16}
\bibfield{author}{\bibinfo{person}{Th{\'{e}}o Trouillon},
  \bibinfo{person}{Johannes Welbl}, \bibinfo{person}{Sebastian Riedel},
  \bibinfo{person}{{\'{E}}ric Gaussier}, {and} \bibinfo{person}{Guillaume
  Bouchard}.} \bibinfo{year}{2016}\natexlab{}.
\newblock \showarticletitle{Complex Embeddings for Simple Link Prediction}.
\newblock \bibinfo{journal}{\emph{Proceedings {ICML}}} (\bibinfo{year}{2016}),
  \bibinfo{pages}{2071--2080}.
\newblock


\bibitem[\protect\citeauthoryear{Wang, Wang, and Guo}{Wang
  et~al\mbox{.}}{2015}]%
        {Rule:conf/ijcai/WangWG15}
\bibfield{author}{\bibinfo{person}{Quan Wang}, \bibinfo{person}{Bin Wang},
  {and} \bibinfo{person}{Li Guo}.} \bibinfo{year}{2015}\natexlab{}.
\newblock \showarticletitle{Knowledge Base Completion Using Embeddings and
  Rules}.
\newblock \bibinfo{journal}{\emph{Proceedings of {IJCAI}}}
  (\bibinfo{year}{2015}), \bibinfo{pages}{1859--1866}.
\newblock


\bibitem[\protect\citeauthoryear{Wang, He, Feng, Nie, and Chua}{Wang
  et~al\mbox{.}}{2018}]%
        {TEM:conf/www/Wang0FNC18}
\bibfield{author}{\bibinfo{person}{Xiang Wang}, \bibinfo{person}{Xiangnan He},
  \bibinfo{person}{Fuli Feng}, \bibinfo{person}{Liqiang Nie}, {and}
  \bibinfo{person}{Tat{-}Seng Chua}.} \bibinfo{year}{2018}\natexlab{}.
\newblock \showarticletitle{{TEM:} Tree-enhanced Embedding Model for
  Explainable Recommendation}. In \bibinfo{booktitle}{\emph{Proceedings of
  {WWW}}}. \bibinfo{pages}{1543--1552}.
\newblock


\bibitem[\protect\citeauthoryear{Wang, Zhang, Feng, and Chen}{Wang
  et~al\mbox{.}}{2014}]%
        {TransH:conf/aaai/WangZFC14}
\bibfield{author}{\bibinfo{person}{Zhen Wang}, \bibinfo{person}{Jianwen Zhang},
  \bibinfo{person}{Jianlin Feng}, {and} \bibinfo{person}{Zheng Chen}.}
  \bibinfo{year}{2014}\natexlab{}.
\newblock \showarticletitle{Knowledge Graph Embedding by Translating on
  Hyperplanes}.
\newblock \bibinfo{journal}{\emph{Proceedings of {AAAI}}}
  (\bibinfo{year}{2014}), \bibinfo{pages}{1112--1119}.
\newblock


\bibitem[\protect\citeauthoryear{Xiao, Huang, and Zhu}{Xiao
  et~al\mbox{.}}{2016}]%
        {TransG:conf/acl/0005HZ16}
\bibfield{author}{\bibinfo{person}{Han Xiao}, \bibinfo{person}{Minlie Huang},
  {and} \bibinfo{person}{Xiaoyan Zhu}.} \bibinfo{year}{2016}\natexlab{}.
\newblock \showarticletitle{TransG : {A} Generative Model for Knowledge Graph
  Embedding}.
\newblock \bibinfo{journal}{\emph{Proceedings of {ACL}}}
  (\bibinfo{year}{2016}).
\newblock


\bibitem[\protect\citeauthoryear{Xie, Liu, and Sun}{Xie et~al\mbox{.}}{2016}]%
        {TKRL:conf/ijcai/XieLS16}
\bibfield{author}{\bibinfo{person}{Ruobing Xie}, \bibinfo{person}{Zhiyuan Liu},
  {and} \bibinfo{person}{Maosong Sun}.} \bibinfo{year}{2016}\natexlab{}.
\newblock \showarticletitle{Representation Learning of Knowledge Graphs with
  Hierarchical Types}.
\newblock \bibinfo{journal}{\emph{Proceedings of {IJCAI}}}
  (\bibinfo{year}{2016}), \bibinfo{pages}{2965--2971}.
\newblock


\bibitem[\protect\citeauthoryear{Yang, tau Yih, He, Gao, and Deng}{Yang
  et~al\mbox{.}}{2015}]%
        {DistMul:conf/iclr/2015}
\bibfield{author}{\bibinfo{person}{Bishan Yang}, \bibinfo{person}{Wen tau Yih},
  \bibinfo{person}{Xiaodong He}, \bibinfo{person}{Jianfeng Gao}, {and}
  \bibinfo{person}{Li Deng}.} \bibinfo{year}{2015}\natexlab{}.
\newblock \showarticletitle{Embedding Entities and Relations for Learning and
  Inference in Knowledge Bases}.
\newblock \bibinfo{journal}{\emph{Proceedings of {ICLR}}}
  (\bibinfo{year}{2015}).
\newblock


\bibitem[\protect\citeauthoryear{Yang, Yang, and Cohen}{Yang
  et~al\mbox{.}}{2017}]%
        {NeurlLP:conf/nips/YangYC17}
\bibfield{author}{\bibinfo{person}{Fan Yang}, \bibinfo{person}{Zhilin Yang},
  {and} \bibinfo{person}{William~W. Cohen}.} \bibinfo{year}{2017}\natexlab{}.
\newblock \showarticletitle{Differentiable Learning of Logical Rules for
  Knowledge Base Reasoning}. In \bibinfo{booktitle}{\emph{{NIPS}}}.
  \bibinfo{pages}{2316--2325}.
\newblock


\bibitem[\protect\citeauthoryear{Yin, Jiang, Lu, Shang, Li, and Li}{Yin
  et~al\mbox{.}}{2016}]%
        {QA:conf/ijcai/YinJLSLL16}
\bibfield{author}{\bibinfo{person}{Jun Yin}, \bibinfo{person}{Xin Jiang},
  \bibinfo{person}{Zhengdong Lu}, \bibinfo{person}{Lifeng Shang},
  \bibinfo{person}{Hang Li}, {and} \bibinfo{person}{Xiaoming Li}.}
  \bibinfo{year}{2016}\natexlab{}.
\newblock \showarticletitle{Neural Generative Question Answering}. In
  \bibinfo{booktitle}{\emph{Proceedings of {IJCAI}}}.
\newblock


\bibitem[\protect\citeauthoryear{Zhang}{Zhang}{2017}]%
        {ORC:conf/www/Zhang17}
\bibfield{author}{\bibinfo{person}{Wen Zhang}.}
  \bibinfo{year}{2017}\natexlab{}.
\newblock \showarticletitle{Knowledge Graph Embedding with Diversity of
  Structures}.
\newblock \bibinfo{journal}{\emph{Proceedings {WWW} Companion}}
  (\bibinfo{year}{2017}), \bibinfo{pages}{747--753}.
\newblock


\end{thebibliography}

\end{document}